\documentclass[10pt,twocolumn,letterpaper]{article}

\usepackage{cvpr}              %

\usepackage[accsupp]{axessibility}  %

\usepackage{lineno}

\definecolor{Gray}{gray}{0.9}
\definecolor{mygreen}{rgb}{0.0, 0.5, 0.0}
\definecolor{myred}{rgb}{0.8, 0.25, 0.33}
\definecolor{myblue}{rgb}{0.19, 0.55, 0.91}
\definecolor{uclablue}{rgb}{0.15, 0.45, 0.68}
\definecolor{ucladblue}{rgb}{0.0, 0.33, 0.53}
\definecolor{ucladdblue}{rgb}{0.0, 0.23, 0.36}
\definecolor{uclagold}{rgb}{1.0, 0.82, 0.0}
\definecolor{ucladgold}{rgb}{1.0, 0.78, 0.17}
\definecolor{ucladdgold}{rgb}{1.0, 0.72, 0.11}
\definecolor{boxgreen}{rgb}{0.02, 0.66, 0.02}
\definecolor{boxred}{rgb}{0.66, 0.1, 0.1}
\definecolor{boxblue}{rgb}{0.01, 0.01, 0.73}

\usepackage{etoc}
\usepackage{lipsum}
\usepackage{wrapfig}
\usepackage{multicol}
\usepackage{mdframed}

\usepackage{mathtools}
\usepackage{amsthm}

\usepackage[utf8]{inputenc}   %
\usepackage[T1]{fontenc}      %
\usepackage{url}              %
\usepackage{amsfonts}         %
\usepackage{nicefrac}         %
\usepackage{microtype}
\usepackage{times}
\usepackage{graphicx}
\usepackage{amsmath,amssymb,mathbbol}
\usepackage{acronym}
\usepackage{enumitem}
\usepackage{balance}
\usepackage{xspace}
\usepackage{setspace}
\usepackage[skip=3pt,font=small]{caption}
\usepackage{booktabs,tabularx,colortbl,multirow,array,makecell}
\usepackage{pifont}
\usepackage{xr-hyper}
\usepackage{tcolorbox}
\usepackage{xcolor}
\tcbuselibrary{breakable}

\definecolor{cvprblue}{rgb}{0.21,0.49,0.74}
\usepackage[pagebackref,breaklinks,colorlinks,allcolors=cvprblue]{hyperref}
\usepackage[capitalise]{cleveref}

\usepackage{siunitx}   %
\usepackage{soul}   %
\usepackage{listings}
\definecolor{codegreen}{rgb}{0,0.6,0}
\definecolor{codegray}{rgb}{0.5,0.5,0.5}
\definecolor{codepurple}{rgb}{0.58,0,0.82}
\definecolor{backcolour}{rgb}{0.95,0.95,0.92}

\lstdefinestyle{mystyle}{
    backgroundcolor=\color{backcolour},   
    commentstyle=\color{codegreen},
    keywordstyle=\color{magenta},
    numberstyle=\tiny\color{codegray},
    stringstyle=\color{codepurple},
    basicstyle=\ttfamily\footnotesize,
    breakatwhitespace=false,         
    breaklines=true,                 
    captionpos=b,                    
    keepspaces=true,                 
    numbers=left,                    
    numbersep=5pt,                  
    showspaces=false,                
    showstringspaces=false,
    showtabs=false,                  
    tabsize=2
}

\makeatletter
\DeclareRobustCommand\onedot{\futurelet\@let@token\@onedot}
\def\@onedot{\ifx\@let@token.\else.\null\fi\xspace}
\def\eg{\emph{e.g}\onedot} 

\def\ie{\emph{i.e}\onedot}

\def\etc{\emph{etc}\onedot} 
\def\vs{\emph{vs}\onedot}

\makeatother

\crefname{algorithm}{Alg.}{Algs.}
\crefname{algocf}{Alg.}{Algs.}
\Crefname{algocf}{Algorithm}{Algorithms}
\crefname{section}{Sec.}{Secs.}
\Crefname{section}{Section}{Sections}
\crefname{table}{Tab.}{Tabs.}
\Crefname{table}{Table}{Tables}
\crefname{figure}{Fig.}{Fig.}
\Crefname{figure}{Figure}{Figure}

\definecolor{gblue}{HTML}{4285F4}
\definecolor{gred}{HTML}{DB4437}
\definecolor{ggreen}{HTML}{0F9D58}

\definecolor{mygray}{gray}{.92}

\newcommand{\supp}{\textit{Appendix}\xspace}

\newcommand{\sota}{state-of-the-art\xspace}

\newcommand{\benchmark}{\textsc{Beacon3D}\xspace}

\acrodef{vl}[VL]{vision-language}
\acrodef{2dvl}[2D-VL]{2D vision-language}
\acrodef{3dvl}[3D-VL]{3D vision-language}
\acrodef{vla}[VLA]{vision-language-action}
\acrodef{qa}[QA]{question answering}
\acrodef{3dqa}[3D-QA]{3D question-answering}
\acrodef{llm}[LLM]{large language model}
\acrodef{vlm}[VLM]{vision-language model}
\acrodef{lvlm}[LVLM]{large vision-language model}
\acrodef{cot}[CoT]{Chain-of-Thought}
\acrodef{ocot}[O-CoT]{Object-centric Chain-of-Thought}
\acrodef{gc}[G-Chain]{Grounding-Chain}
\acrodef{gqac}[GQA-Chain]{Grounding-QA-Chain}

\definecolor{color1}{rgb}{0.81,0.77,0.71}
\definecolor{color2}{rgb}{0.92,1.0,0.98}
\definecolor{color3}{rgb}{0.89,0.90,0.98}
\definecolor{color4}{rgb}{0.9,0.96,1.0}

\usepackage{amsmath,amsfonts,bm}

\def\eqref#1{equation~\ref{#1}}

\def\1{\bm{1}}

\def\vs{{\bm{s}}}

\DeclareMathAlphabet{\mathsfit}{\encodingdefault}{\sfdefault}{m}{sl}
\SetMathAlphabet{\mathsfit}{bold}{\encodingdefault}{\sfdefault}{bx}{n}

\def\paperID{5910} %
\def\confName{CVPR}
\def\confYear{2025}

\title{Unveiling the Mist over 3D Vision-Language Understanding:\\Object-centric Evaluation with Chain-of-Analysis}

\author{Jiangyong Huang$^{1,2,*}$ \quad Baoxiong Jia$^{1,*}$ \quad Yan Wang$^1$ \quad Ziyu Zhu$^{1,3}$ \quad Xiongkun Linghu$^1$ \\ Qing Li$^1$ \quad Song-Chun Zhu$^{1,2,3}$ \quad Siyuan Huang$^{1}$\vspace{0.2em}\\
{\small $^1$State Key Laboratory of General Artificial Intelligence, BIGAI}\vspace{-0.1em}\\
{\small $^2$Peking University, $^3$Tsinghua University
}\vspace{0.1em}\\
\href{https://beacon-3d.github.io}{https://beacon-3d.github.io}\vspace{-0.5em}
}

\begin{document}
\twocolumn[{
\renewcommand\twocolumn[1][]{#1}%
\maketitle
\begin{center}
    \centering
    \vspace{-15pt}
    \captionsetup{type=figure}
    \includegraphics[width=0.97\linewidth]{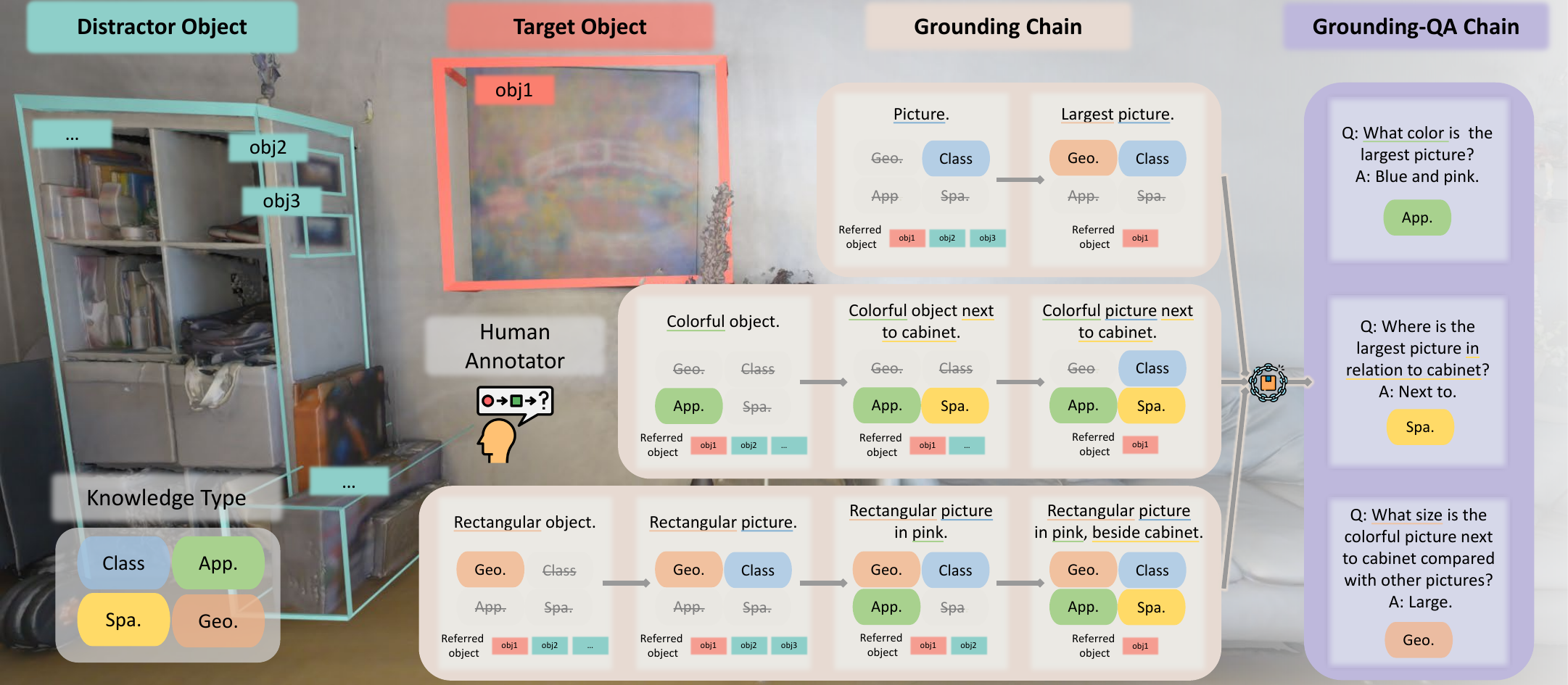}
    \captionof{figure}{\textbf{An overview of \benchmark, a novel benchmark for 3D grounding and question answering (QA) tasks.} \benchmark features an object-centric evaluation framework, with Grounding-Chains (G-Chains) and Grounding-QA-Chains (GQA-Chains) for each object. The evaluation adopts object-centric metrics to ensure robustness and utilizes chain-of-analysis for studies in task coherence. We also involve the study of various knowledge types such as class, appearance (``App.''), spatial (``Spa.''), and geometry (``Geo.'').}
    \label{fig:teaser}
\end{center}%
}]
\let\thefootnote\relax\footnote{$^*$Equal contribution.}

\def\vs{\emph{vs}\onedot}
\begin{abstract}
Existing \ac{3dvl} benchmarks fall short in evaluating \ac{3dvl} models, creating a ``mist'' that obscures rigorous insights into model capabilities and \ac{3dvl} tasks. This mist persists due to three key limitations. First, flawed test data, like ambiguous referential text in the grounding task, can yield incorrect and unreliable test results. Second, oversimplified metrics such as simply averaging accuracy per \ac{qa} pair, cannot reveal true model capability due to their vulnerability to language variations. Third, existing benchmarks isolate the grounding and \ac{qa} tasks, disregarding the underlying coherence that \ac{qa} should be based on solid grounding capabilities. To unveil the ``mist'', we propose \benchmark, a benchmark for \ac{3dvl} grounding and \ac{qa} tasks, delivering a perspective shift in the evaluation of \ac{3dvl} understanding. \benchmark features (i) high-quality test data with precise and natural language, (ii) object-centric evaluation with multiple tests per object to ensure robustness, and (iii) a novel chain-of-analysis paradigm to address language robustness and model performance coherence across grounding and \ac{qa}. Our evaluation of \sota \ac{3dvl} models on \benchmark reveals that (i) object-centric evaluation elicits true model performance and particularly weak generalization in \ac{qa}; (ii) grounding-\ac{qa} coherence remains fragile in current \ac{3dvl} models, and (iii) incorporating \acp{llm} to \ac{3dvl} models, though as a prevalent practice, hinders grounding capabilities and has yet to elevate \ac{qa} capabilities. We hope \benchmark and our comprehensive analysis could benefit the \ac{3dvl} community towards faithful developments.
\end{abstract}
    
\section{Introduction}

The ability to understand 3D scenes is an essential facet of human-level intelligence \citep{marr2010vision,qi2018scene,huang2018holistic,chen2019holistic++,zhu2020dark}. Recent \acl{3dvl} (\ac{3dvl}) models have achieved notable progress in language-grounded 3D scene understanding \citep{chen2022language,zhu20233d,gong2023arnold,hong20233dllm,huang2024embodied,chen2024ll3da,jia2024sceneverse,zhu2024unifying,huang2024chatscene,ma2024llms}, and various benchmarks have been established for \ac{3dvl} tasks like object grounding \citep{chen2020scanrefer,achlioptas2020referit3d,zhang2023multi3drefer,wang2024embodiedscan,jia2024sceneverse,yang20243d} and \acl{qa} (\ac{qa}) \citep{azuma2022scanqa,ma2023sqa3d,hong20233dqa,majumdar2024openeqa}. Despite the improving performance on these benchmarks, a critical question remains to be addressed:

\textit{How effective are these benchmarks for \ac{3dvl} understanding; are the progress and results on these benchmarks reliable enough to guide the development of \ac{3dvl} models?}

We raise considerable concerns on this question, observing several key limitations in existing \ac{3dvl} benchmarks: 
\begin{itemize}[leftmargin=*,nolistsep,noitemsep]
    \item First, we observe notable flaws in the test data, which may undermine the reliability of evaluations. For example, referential text in the grounding task can be \textit{ambiguous} or \textit{unnatural}, leading to ill-posed tests; \textit{ambiguous questions} in \ac{qa} data may mislead to divergent answers; \textit{incomplete answer labels} can misrepresent model performance by penalizing correct predictions. Our human studies highlight these flaws in ScanRefer \citep{chen2020scanrefer} and ScanQA \citep{azuma2022scanqa}, as validated by the limited human performance. Additionally, we show that addressing the flaws in ScanRefer can lead to a more accurate evaluation of model performance.
    
    \item Second, the evaluation metrics in current \ac{3dvl} benchmarks fall short in accurately capturing model capability. Oversimplified metrics, such as averaging accuracy over individual \ac{qa} pairs, are vulnerable to model pitfalls like \textit{visual ignorance} (\ie, predictions determined solely by texts) and \textit{weak language robustness} (\ie, predictions susceptible to varied texts). We demonstrate their vulnerability by showing that blind \acp{llm} can achieve unexpectedly high accuracy on SQA3D \citep{ma2023sqa3d}, and even minor language rephrasing can significantly affect \ac{qa} accuracy. This suggests the need for more robust evaluation metrics through language variations and multiple tests for each object.
    
    \item Third, current \ac{3dvl} benchmarks isolate grounding and \ac{qa} tasks, exposing \ac{qa} in the risk of shortcuts. To address this gap, we design \acp{gqac} to assess model performance coherence between grounding and \ac{qa}. These chains ensure that the contents of \ac{qa} are covered by corresponding grounding texts. Our study on \acp{gqac} reveals two types of broken coherence: (i) \textit{model correctly grounds the object but fails in \ac{qa}}, showing poor \ac{qa} skills; and (ii) \textit{model fails in grounding but succeeds in \ac{qa}}, suggesting shortcuts in \ac{qa}. Specifically, on a \sota \ac{3dvl} model PQ3D \citep{zhu2024unifying}, we observe that half of \ac{qa} errors are associated with correct grounding predictions, while one-quarter of correct answers result from shortcuts. This implies the potentially fragile grounding-\ac{qa} coherence in \ac{3dvl} models.
\end{itemize}

Motivated by our analyses, we construct \benchmark, a novel benchmark for \ac{3dvl} grounding and \ac{qa} tasks, providing a new perspective in \ac{3dvl} evaluation. The benchmark is built on 30 meticulously selected high-quality scenes from ScanNet \citep{dai2017scannet}, 3RScan \citep{wald2019rio}, and MultiScan \citep{mao2022multiscan}. We exhaustively annotate objects in each scene and introduce object-level evaluation with three cases per object for both grounding and \ac{qa}. This yields more robust and reliable object-centric metrics, reflecting the true model capabilities. Additionally, we propose \acp{gc} for the grounding task, spanning grounding texts from coarse (\eg, ``chair'') to fine-grained (\eg, ``gray chair next to the corner table'') descriptions. To address the isolation of grounding and \ac{qa} tasks, we further construct \acp{gqac} associated with \acp{gc} to assess model performance coherence across grounding and \ac{qa} tasks.
\benchmark comprises a total of 837 objects, 2511 \acp{gc} and 2511 \acp{gqac}, with all annotations manually crafted for language clarity and naturalness. 
We employ object-centric evaluation metrics that require accurate predictions across all three tests per object for grounding and \ac{qa}, helping to better manifest model pitfalls. The \acp{gc} and \acp{gqac} also enable a novel chain-of-analysis evaluation paradigm in \benchmark, providing a holistic assessment of \ac{3dvl} model capabilities.

We apply \benchmark to evaluate \sota \ac{3dvl} models. Compared to conventional per-case averages, object-centric metrics elicit a significant model performance drop in both grounding and \ac{qa}. This highlights that models are prone to language variations and exhibit a limited object-level understanding. Analyses on \acp{gc} show that models struggle when the granularity of grounding texts increases. And analyses on \acp{gqac} reveal a fragile grounding-\ac{qa} coherence in \ac{3dvl} models, underscoring the gap between grounding and \ac{qa} skills, and the prevalence of shortcuts in 3D \ac{qa}. Furthermore, contrary to existing practices~\cite{hong20233dllm,huang2024embodied,chen2024ll3da,qi2024shapellm}, our results show that incorporating \acp{llm} for \ac{3dvl} models hinders grounding and has yet to improve \ac{qa} performance on \benchmark, offering new insights into the learning of grounding and \ac{qa} tasks. 

We summarize our contributions as follows:
\begin{enumerate}[noitemsep]
    \item We present detailed investigations into limitations of existing \ac{3dvl} benchmarks and expose fragile performance coherence across grounding and \ac{qa} in \ac{3dvl} models.
    \item We propose \benchmark, a benchmark for 3D grounding and \ac{qa} that shifts the evaluation paradigm to object-centric evaluation with chain-of-analysis on grounding and grounding-\ac{qa} chains, providing a high-quality, faithful, and holistic tool for evaluating \ac{3dvl} models.
    \item We present a comprehensive analysis of \sota \ac{3dvl} models on \benchmark, highlighting common model pitfalls like grounding-\ac{qa} incoherence and incomplete object understanding, along with the unexpected hindrance of LLM for \ac{3dvl} tasks.
\end{enumerate}

\begin{figure*}[t!]
    \centering
    \includegraphics[width=\linewidth]{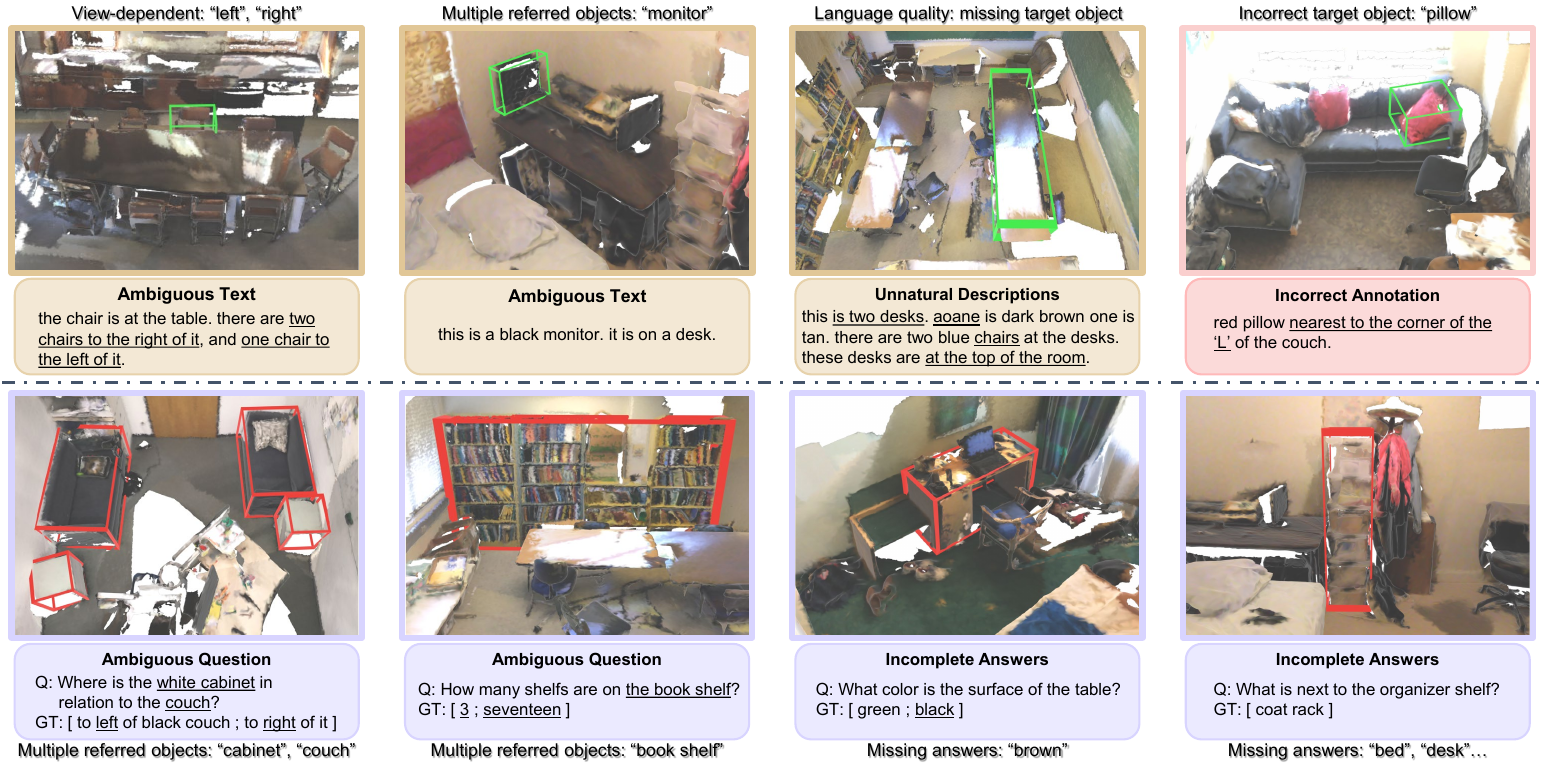}
    \caption{\textbf{Various types of test data flaws in \textcolor{Dandelion}{ScanRefer}, \textcolor{Lavender}{Nr3D}, \textcolor{Orchid}{ScanQA}.} \underline{Underlined texts} indicate explicit flaws. (1) The top row shows grounding data with the target object highlighted. \textbf{Ambiguous text} includes viewpoint-dependent expressions like ``left'' and ``right'', or lacks information to uniquely specify the target object. \textbf{Unnatural descriptions} are hard to understand by humans for being too tedious or grammatically invalid. \textbf{Incorrect annotation} refers to the mismatch between text and target object. (2) The bottom row shows \ac{qa} data with ground truth (GT) shown in square brackets. \textbf{Ambiguous question} lacks context to clarify the queried object, potentially leading to contradictory answers. \textbf{Incomplete answers} may forbid alternative correct answers.}
    \label{fig:data_flaw}
    \vspace{-12pt}
\end{figure*}

\section{Related Work}

\paragraph{\acl{3dvl} models.} Fueled by the advancement of \acp{vlm} \citep{yu2019deep,radford2021learning,li2022language,ghiasi2022scaling,huang2022perceive,kirillov2023segment,openai2023gpt4} and reconstruction techniques \citep{wang2024roomtex,lu2025movis,wan2024superpoint,chen2024ssr,ni2024phyrecon,ni2025dprecon,liu2025building,liu2024slotlifter,jia2023improving,yu2025metascenes}, the capability of 3D scene understanding has been greatly improved. Key contributions in this area include 3D perception techniques \citep{qi2017pointnet,qi2017pointnet++,zhao20213dvg,abdelreheem20223dreftransformer,chen2022language,huang2022multi,schult2023mask3d,lu2025movis,wang2025masked}, 2D-3D feature integration \citep{yang2021sat,ha2022semantic,jatavallabhula2023conceptfusion,peng2023openscene,kerr2023lerf,zhu2024unifying}, and \ac{3dvl} pretraining \citep{zhu20233d,ding2023pla,zhou2024uni3d,xue2024ulip,jia2024sceneverse,wang2024embodiedscan}. On the other hand, the rapid development of \acp{lvlm} \citep{li2023blip2,liu2023visual,dai2023instructblip} drives \ac{3dvl} models to evolve from task-specific architectures to generalist frameworks \citep{hong20233dllm,huang2024embodied,yang2024llm,chen2024ll3da,fu2024scene,zhang2024task,chu2024unified,huang2024chatscene,zhu2024llava,kang2024robin3d}. While these 3D \acp{lvlm} demonstrate impressive capabilities, there is also a pressing demand for advanced benchmarks to comprehensively evaluate these models, and address underexplored questions, \eg, generalizability and the effect of 
\acp{llm}.

\paragraph{3D \acl{vl} datasets and benchmarks.} Early research in \ac{3dvl} learning has produced initial task-specific benchmarks for grounding \citep{chen2020scanrefer,achlioptas2020referit3d,zhang2023multi3drefer} and \ac{qa} \citep{azuma2022scanqa,ye20223d,ma2023sqa3d,hong20233dqa}, akin to the early stage of \ac{2dvl} benchmarks \citep{yu2016modeling,mao2016generation,antol2015vqa,hudson2019gqa,marino2019ok,schwenk2022okvqa}. As recent \acp{lvlm} evolve to be more powerful and intricate, 2D \ac{vl} benchmarks have advanced towards meticulously designed evaluation or detailed analysis \citep{yuksekgonul2022and,liu2023visual,fu2023mme,yu2023mm,yue2024mmmu,tong2024eyes,rahmanzadehgervi2024vision,li2023evaluating,liu2023mmbench,chen2024we}. In contrast, recent \ac{3dvl} works mainly focus on large-scale learning \citep{luo2023scalable,zhu20233d,huang2024embodied,wang2024embodiedscan,jia2024sceneverse,linghu2024multi,lyu2024mmscan} while adhering to conventional evaluation criteria \citep{chen2020scanrefer,achlioptas2020referit3d,azuma2022scanqa,ma2023sqa3d}. On the other hand, recent advance in the evaluation of \ac{3dvl} models \citep{singh2024evaluating,majumdar2024openeqa,szymanska2024space3d,man2024lexicon3d,straub2024efm3d,zhao2024openscan} provides suites for analyzing issues such as hallucination and robustness \citep{yang20243d,kang2024robin3d,deng2024can}. Nonetheless, prior works have not established an evaluation criterion with reliable metrics and in-depth analysis of 3D grounding and \ac{qa} tasks, which is the exact goal of this paper.

\section{An Investigation into \ac{3dvl} Benchmarks}\label{sec:3}

\subsection{Flawed Test Data}\label{sec:3.1}

When examining existing \ac{3dvl} benchmarks, we identified flaws in the test data as a significant issue for evaluating model performance. We provide justifications from both quantitative and qualitative aspects as follows:

\begin{table*}[t!]
\begin{minipage}{0.49\linewidth}
    \centering
    \caption{\textbf{Human study on ScanRefer val set.} We report clarity and naturalness scores (1$\sim$5) of the referential text, as well as human and model prediction accuracy. We use PQ3D~\cite{zhu2024unifying} for model evaluation.}
    \resizebox{\linewidth}{!}{
    \begin{tabular}{c|cccc}
        \toprule
        Data Source & Clarity & Naturalness & Human Accuracy & Model Accuracy \\
        \midrule
        ScanRefer & 3.70       & 4.23     & 69\%                & 63\% \\
        Refined   & 4.59       & 4.34   & \textbf{100\%}                & \textbf{70\%} \\
        \bottomrule
    \end{tabular}
    }
    \label{tab:scanrefer_human_study}
\end{minipage}
\hfill
\begin{minipage}{0.47\linewidth}
    \centering
    \caption{\textbf{Human study on ScanQA (val) and SQA3D (val and test).} Quality scores range from 1 to 5. Human accuracy is evaluated using answer labels as the ground truth.}
    \resizebox{\linewidth}{!}{
    \begin{tabular}{c|ccc}
        \toprule
        Data Source & Question Quality & Answer Quality & Human Accuracy \\
        \midrule
        ScanQA & 3.44       & 3.60     & 62\% \\
        SQA3D   & 4.64       & 4.46   & 80\% \\
        \bottomrule
    \end{tabular}
    }
    \label{tab:scanqa_human_study}
\end{minipage}
\vspace{-5pt}
\end{table*}
\begin{figure*}[t!]
\begin{minipage}{0.49\linewidth}
    \centering
    \includegraphics[width=\linewidth]{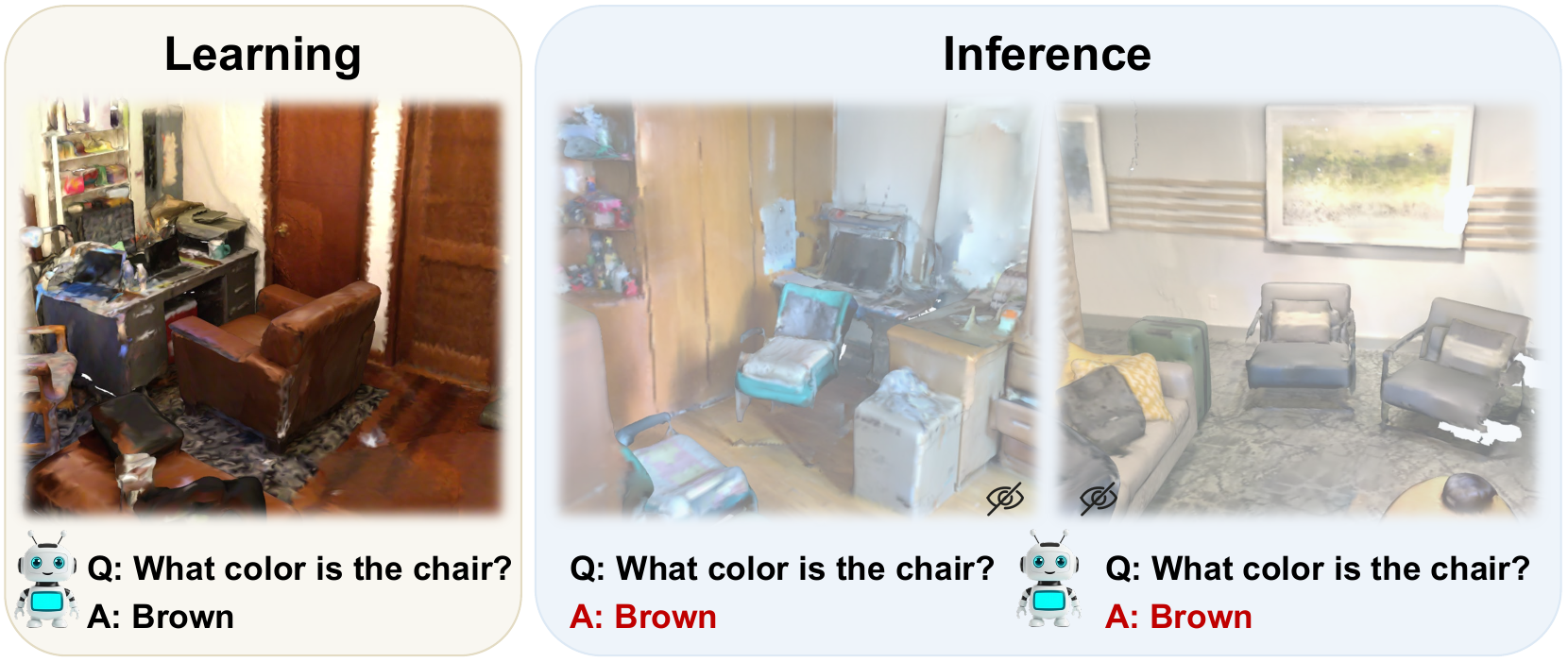}
    \caption{\textbf{Illustrative examples on visual ignorance.} The model predicts answers directly from questions, ignoring scene information (\eg, chair color).}
    \label{fig:visual_ignorance}
\end{minipage} 
\hfill
\begin{minipage}{0.49\linewidth}
    \centering
    \includegraphics[width=\linewidth]{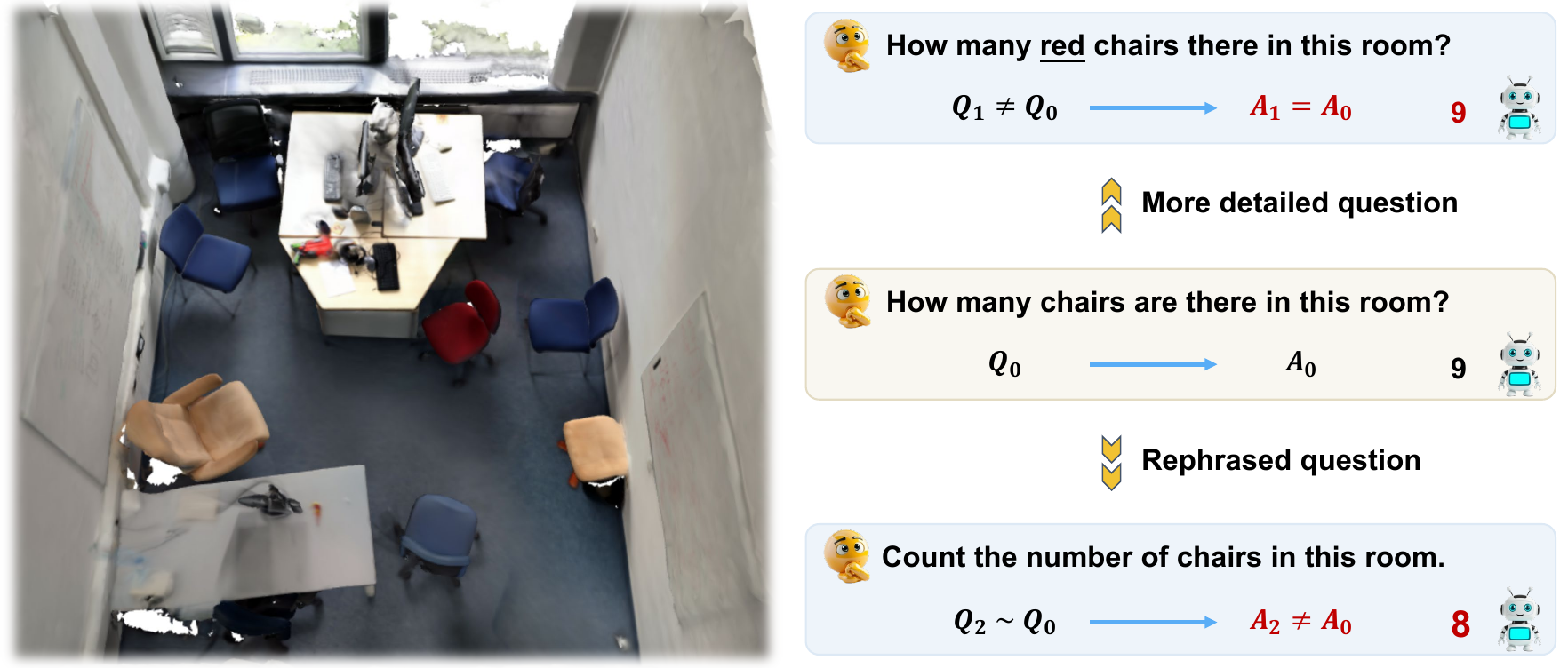}
    \caption{\textbf{Illustrative examples on language robustness.} Rephrased and more detailed questions of the same concept can easily lead to wrong model predictions.}
    \label{fig:language_robustness}
\end{minipage}
\vspace{-12pt}
\end{figure*}

\paragraph{Qualitative analysis.} We analyze the test data quality from prevalent \ac{3dvl} benchmarks: ScanRefer \citep{chen2020scanrefer} and Nr3D \citep{achlioptas2020referit3d} for grounding, and ScanQA \citep{azuma2022scanqa} for 3D-\ac{qa}. We identify common data flaws, shown in \cref{fig:data_flaw}. Key grounding issues include: (i) \textit{ambiguous referential text}, which lacks information to uniquely identify the target object; and (ii) \textit{unnatural descriptions}, being excessively complex, that are difficult to identify the target object. For 3D-\ac{qa}, we observe that (i) \textit{ambiguous questions} with no clear targeting object easily leads to contradictory answers, and (ii) questions with \textit{incomplete answers} can undermine evaluation reliability by forbidding alternative valid answers predicted by the models.

\paragraph{Quantitative analysis.} We provide quantitative measurements of data flaws and their impacts. For grounding, we sample a subset of 100 grounding texts from the ScanRefer validation set and instruct human evaluators to re-predict the target object based on the referential text and score the \textit{clarity} and \textit{naturalness} of each text (scored from 1 to 5). As shown in~\cref{tab:scanrefer_human_study}, a large portion (31\%) of the test data leads to incorrect human predictions. We test a recent \sota \ac{3dvl} model, PQ3D~\cite{zhu2024unifying}, before and after manually refining these texts. We observe a significant model performance improvement (7\%) without model-side adjustments. 

For \ac{qa}, we also randomly sample 100 QA pairs from ScanQA and SQA3D~\cite{ma2023sqa3d}. We instruct human evaluators to re-answer the questions and rate the quality of the \ac{qa} text. As shown in~\cref{tab:scanqa_human_study}, the low human prediction accuracy (62\% on ScanQA) highlights that \textit{the flaws in \ac{qa} data pose a tangible upper bound on model performance}. These analyses on existing grounding and \ac{qa} benchmark underscore the need for rigorous quality control in \ac{3dvl} benchmarks.

\subsection{Insufficient Evaluation Metrics}\label{sec:3.2}

\begin{figure*}[t!]
    \centering
    \includegraphics[width=\linewidth]{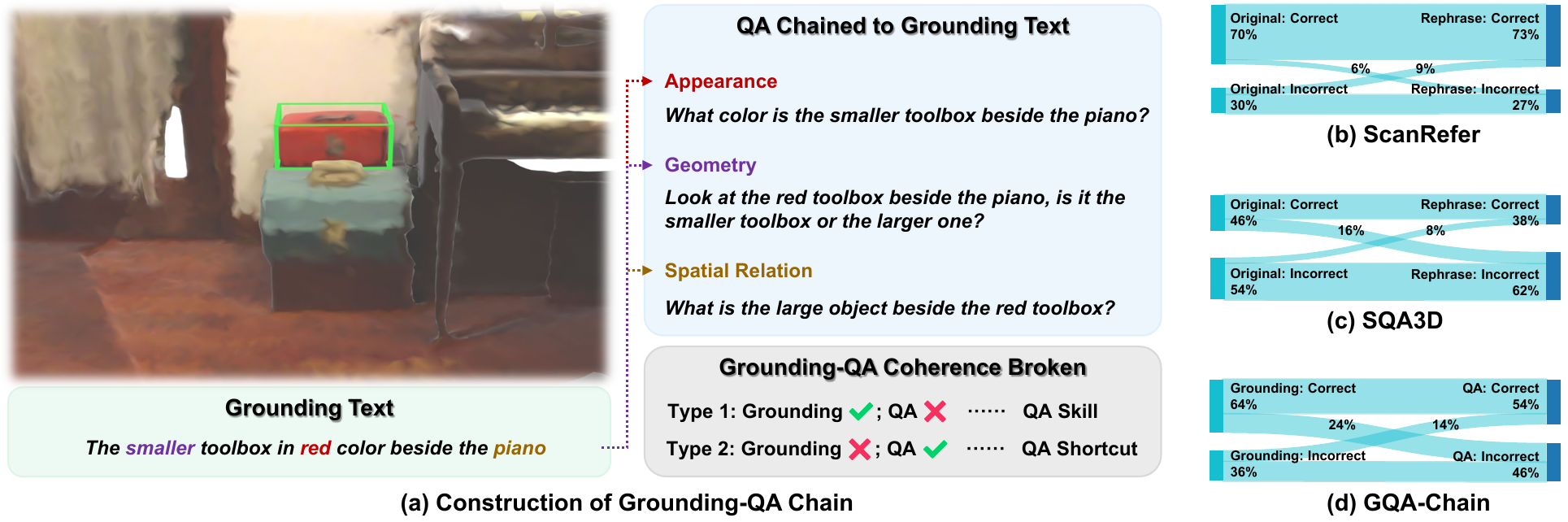}
    \captionof{figure}{\textbf{(a) Illustration of \acp{gqac}.} The questions derive from the grounding text and query a specific feature of the target object. We define two broken types for grounding-\ac{qa} coherence: (Type 1) correct grounding and incorrect \ac{qa}, indicating a lack of \ac{qa} skills; (Type 2) incorrect grounding and correct \ac{qa}, suggesting shortcuts in \ac{qa}. \textbf{(b) The effect of rephrasing ScanRefer texts on the performance of PQ3D.} \textbf{(c) The effect of rephrasing SQA3D questions on the performance of PQ3D.} \textbf{(d) Results of PQ3D on \acp{gqac}.} We observe over half of \ac{qa} failures (24\% out of 46\%) stem from insufficient \ac{qa} skills while nearly a quarter of correct QA predictions (14\% out of 54\%) are achieved via shortcuts.}
    \label{fig:sankey}
    \vspace{-12pt}
\end{figure*}

In this section, we show that simple metrics like average accuracy over all test instances in existing \ac{3dvl} benchmarks are insufficient to reveal true model pitfalls including \textit{visual ignorance} and poor \textit{language robustness}:
\begin{itemize}[leftmargin=*,nolistsep,noitemsep]
    \item\textbf{Visual ignorance} refers to the scenario where models can perform tasks without the need for visual input, as illustrated in \cref{fig:visual_ignorance}. As an example, we show in~\cref{tab:blind_llm} that fine-tuning ``blind'' \acp{llm} yields a comparable result on SQA3D metrics compared to \sota \ac{3dvl} models. This indicates a deficiency in SQA3D's metrics for evaluating the visual capability of \ac{3dvl} models.
    
    \item\textbf{Language robustness} refers to a model's susceptibility to language variations. For example, in \ac{qa} (see \cref{fig:language_robustness}), models often struggle with \textit{rephrased} or \textit{more detailed} questions about the same object concept (\eg, chairs). We demonstrate this by rephrasing good questions sampled in~\cref{sec:3.2} and comparing PQ3D's performance on the rephrased sets versus the original sets. The results in~\cref{fig:sankey}(b,c) show model sensitivity to language variations do exist, especially on SQA3D where 16\% of predictions switch from correct to incorrect. However, such a problem is overlooked with current \ac{3dvl} benchmarks treating these variations as separate instances during evaluation.
\end{itemize}

\begin{table}[t!]
    \centering
    \caption{\textbf{Blind \acp{llm} finetuned with LoRA on SQA3D.} $^\dagger$ indicates the performance of \sota \ac{3dvl} model \citep{zhu2024llava}.}
    \resizebox{\linewidth}{!}{
    \begin{tabular}{l|ccccc}
        \toprule
        Blind \ac{llm} & OPT-1.3B & Gemma2-2B & Vicuna-7B & LLaMA3-3B & LLaVA-3D$^\dagger$ \\
        \midrule
        EM-1 & 43.9 & 48.8 & 49.4 & 50.0 & 55.6 \\
        \bottomrule
    \end{tabular}
    }
    \vspace{-10pt}
    \label{tab:blind_llm}
\end{table}

To prevent lingual shortcuts arising from \textit{visual ignorance}, we need careful data curation to avoid scene-irrelevant questions and introduce vision-oriented metrics to assess models' visual capability. To better evaluate \textit{language robustness} of models, we need robust evaluation frameworks that incorporate language variations and multiple evaluation instances per object. Thus, we argue that \ac{3dvl} benchmarks must evolve to better visualize these crucial dimensions of \ac{3dvl} model performance.

\subsection{Grounding-\ac{qa} Coherence}\label{sec:3.3}
During our exploration, one critical question we identified, yet has been overlooked by existing benchmarks, is: \textit{Why do models fail in \acs{3dqa} tasks; is it due to language complexity or inadequate scene understanding capabilities?} Believing that accurate \ac{qa} predictions should be grounded in strong scene understanding, we propose a novel \acl{gqac} (\acs{gqac}) that connects grounding and \ac{qa} evaluations to provide detailed analyses of model performance coherence across tasks. The core idea behind \acp{gqac} is to align questions with referential descriptions, ensuring the queried content is directly present in the descriptive texts. For example, in~\cref{fig:sankey}(a), the questions ask about the appearance, geometry, and spatial relationships of the target object, all of which are explicitly described in the referential texts. 

With the expectation that strong \ac{3dvl} should exhibit consistent performance across grounding-\ac{qa} pairs in \acp{gqac}, we generate \acp{gqac} based on the refined ScanRefer subset from~\cref{sec:3.1} as a preliminary experiment. We evaluate PQ3D on both datasets and visualize the results in \cref{fig:sankey}(d). We observe that \textbf{over half of \ac{qa} failures stem from insufficient \ac{qa} skills while nearly a quarter of correct \ac{qa} predictions are achieved via shortcuts}. These findings suggest the prevalence of broken grounding-\ac{qa} coherence in 3D \ac{vl} models, as well as the demand for benchmarks to systematically evaluate grounding-\ac{qa} coherence.

\section{The \benchmark Benchmark}\label{sec:4}

In this section, we introduce \benchmark, a novel benchmark for \ac{3dvl} grounding and \ac{qa} tasks that addresses key evaluation limitations identified in~\cref{sec:3}. We propose the formats of \acl{gc} (\acs{gc}) and \acl{gqac} (\acs{gqac}) for organizing grounding and QA data, along with an object-centric chain-of-analysis paradigm that evaluates models' performance coherence under language variations and across tasks using object-centric metrics.

\begin{figure*}[t!]
\begin{minipage}{0.35\linewidth}
    \centering
    \includegraphics[width=\linewidth]{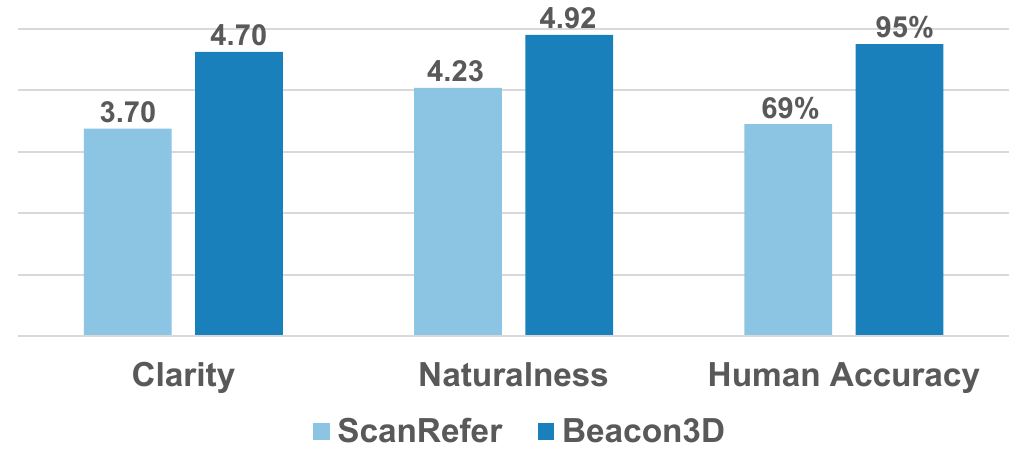}
    \caption{\textbf{Human study on grounding data.}}
    \label{fig:quality_refer}

    \centering
    \includegraphics[width=\linewidth]{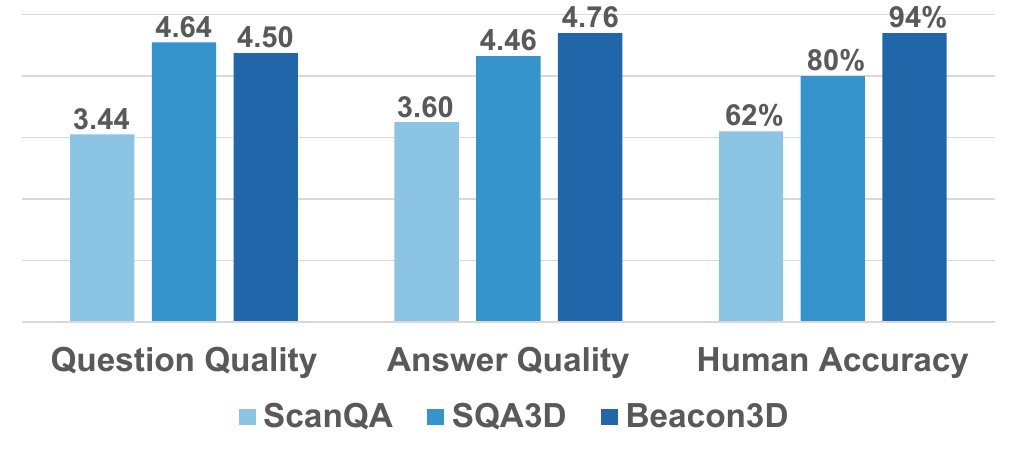}
    \caption{\textbf{Human study on \ac{qa} data.}}
    \label{fig:quality_qa}
\end{minipage}
\hfill
\begin{minipage}{0.64\linewidth}
    \centering
    \includegraphics[width=\linewidth]{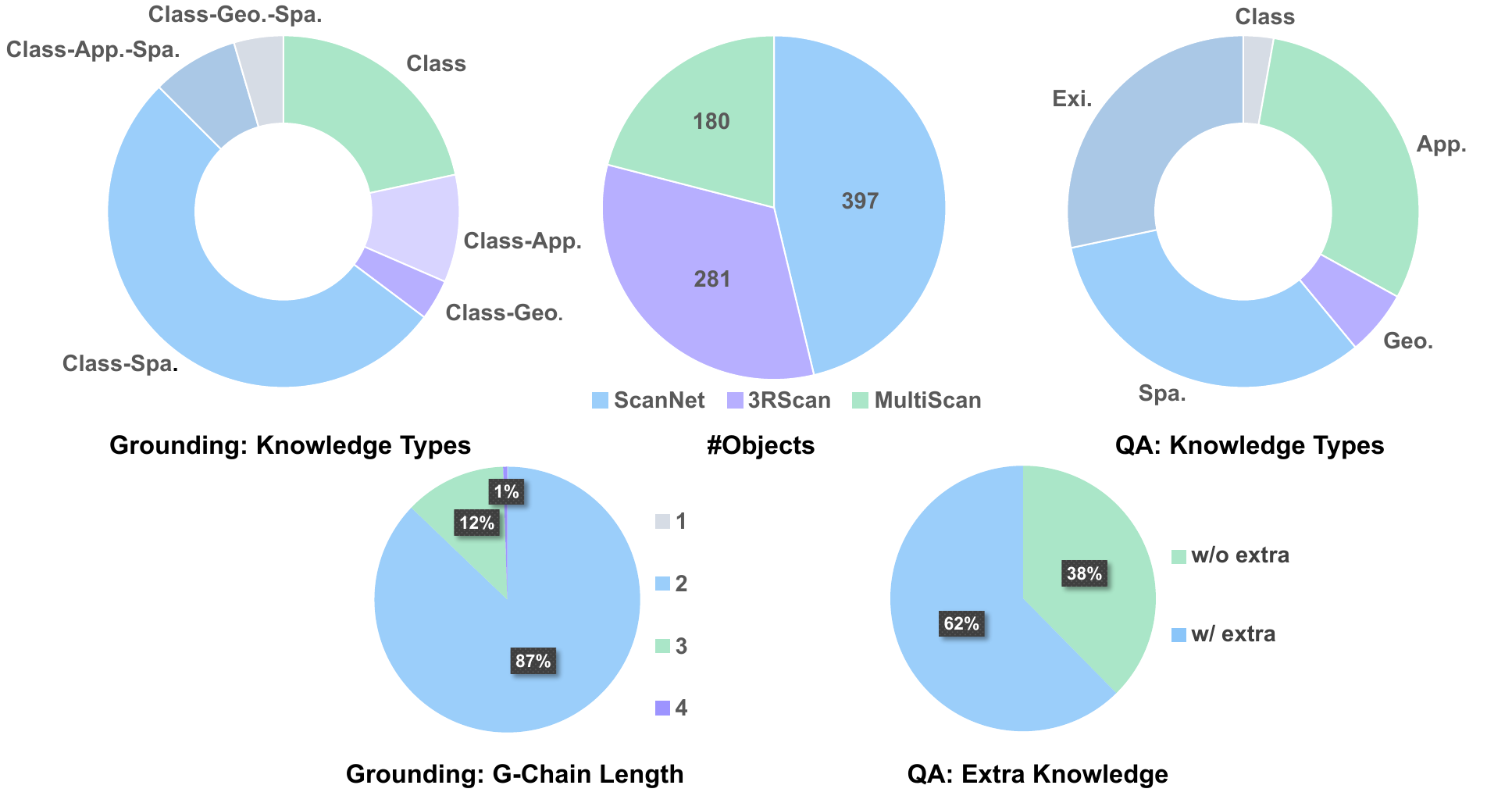}
    \caption{\textbf{Data statistics in \benchmark.}}
    \label{fig:data_stat}
\end{minipage}
\vspace{-10pt}
\end{figure*}

\subsection{Benchmark Design}
\paragraph{Data Design} We consider two tasks in \benchmark: (i) 3D grounding, where models are required to predict the target object's 3D bounding box given the scene point cloud and object referential texts; and (ii) \acs{3dqa}, where models are required answer a question about a target object based on the scene point cloud. The data for these two tasks consists of:
\begin{itemize}[leftmargin=*,nolistsep,noitemsep]

\item \textbf{Grounding:} we create \ac{gc} that consists of a series of referential texts, ranging from coarse to fine. At its finest level, the primary grounding text uniquely identifies the target object. It is then rephrased into progressively coarser texts at each subsequent level, referred to as simplified grounding texts (see in~\cref{fig:teaser}). This relaxation in object descriptions expands the set of correct objects for simplified ground texts at each level, requiring model predictions to fall within its set for correctness evaluation.

\item \textbf{Question Answering:} As in~\cref{sec:3.3}, we construct \acp{gqac} by designing \ac{qa} pairs based on the primary grounding texts in \acp{gc}. Each answer in a \ac{gqac} question is explicitly present in the corresponding primary grounding text. To provide a holistic evaluation, similar to other benchmarks, and accommodate questions that require commonsense knowledge, we also curate a set of questions with queried content not explicitly found in the primary grounding texts. We tag these questions with an \textit{``extra knowledge''} flag and exclude them from the coherence analysis.

\end{itemize}

In addition, we tag each grounding and \ac{qa} data with its required knowledge types: \texttt{class} (semantic category), \texttt{appearance} (color, material, texture, \etc), \texttt{geometry} (shape, size, \etc), and \texttt{spatial-relation}. An extra knowledge type \texttt{exist} is added to \ac{qa} for the questions about whether something exists. Each \ac{qa} data is assigned a single knowledge type according to its \textit{queried content}.

\paragraph{Data Collection} We begin data collection by selecting high-quality scenes from the held-out sets of ScanNet \citep{dai2017scannet}, 3RScan \citep{wald2019rio}, and MultiScan \citep{mao2022multiscan} following two principles: (1) the layout should be reasonable, neither overly cluttered nor too simple, with clear object mesh reconstructions; and (2) objects should be well-placed in the scene with balanced distribution over categories. This results in 30 high-quality scenes in diverse styles from the three datasets. Next, we identify potential target objects by excluding: (i) background objects like walls and floors, (ii) objects that are difficult to distinguish via text (\eg, multiple chairs around a table), and (iii) objects with comparatively low-quality reconstructions, resulting in 837 unique target object instances. We then build an annotation tool following~\cite{ma2023sqa3d} (see details in the \supp) for human annotators to annotate three \acp{gc} and \acp{gqac} for each object instance, totaling 2511 \acp{gc} and 2511 \acp{gqac}. To address prior data flaws, we establish detailed annotation guidelines, ensuring precise and natural language, the indispensability of visual modality in \ac{qa}, and also balanced answer distributions. Each annotation is cross-validated by two human reviewers.

\paragraph{Metrics} In addition to the conventional per-case average metrics, we adopt an object-centric evaluation scheme, requiring models to accurately predict over \textbf{all three} grounding or \ac{qa} test cases per object. Our task-specific metrics are computed as follows:

\begin{itemize}[leftmargin=*,nolistsep,noitemsep]

\item \textbf{Grounding:} For each grounding text, the model is considered correct if the predicted object is included within the candidate object set. For the object-centric metrics, we first derive per-object results according to whether \textbf{all three} predictions on the \textbf{primary grounding texts} are correct, and then average the results over all objects. We also report per-case metrics by averaging the results over all \textbf{primary and simplified grounding texts}.

\item \textbf{Question Answering:} We first evaluate each \ac{qa} pair using GPT-Score \citep{majumdar2024openeqa}, yielding a score $M$ between 1 to 5 from GPT-4 \citep{openai2023gpt4}. The corresponding per-case accuracy is then calculated as \textbf{$\frac{M-1}{4}$} following~\cite{majumdar2024openeqa}. We derive a binary per-object accuracy if \textbf{$M\ge 4$} for \textbf{all three} \ac{qa} pairs. We report object-centric metrics by averaging per-object accuracies, as well as per-case average accuracy over all \textbf{individual \ac{qa} pairs}.

\end{itemize}

\subsection{Data Quality Check and Statistics}
To assess the quality of the data collected in \benchmark, we have a separate group of human annotators evaluate it based on clarity, naturalness, and human accuracy, following metrics used in~\cref{sec:3.1}. For a fair comparison, we sample the same quantity of data from the same scenes. As shown in~\cref{fig:quality_refer,fig:quality_qa}, \benchmark significantly outperforms existing 3D grounding and \ac{qa} benchmarks in terms of language clarity, naturalness, and especially human accuracy metric where nearly $\sim$95\% of the data labeled as correct upon re-examination. We also visualize the statistics of \benchmark in \cref{fig:data_stat}, including object counts by domains, knowledge types, data counts by knowledge types, and the proportion of \ac{qa} pairs requiring \textit{extra knowledge}.

\begin{table*}[t!]
\begin{minipage}{0.468\linewidth}
    \centering
    \captionof{table}{\textbf{Evaluation results of grounding on \benchmark.} The ``Obj.'' column reports object-centric metrics. The columns of knowledge types report per-case averages over each type.}
    \resizebox{\linewidth}{!}{
    \begin{tabular}{lcccccc}
    \toprule
     & \multicolumn{4}{c}{Knowledge type} & \multicolumn{2}{c}{Overall} \\
     \cmidrule(lr){2-5} \cmidrule{6-6} \cmidrule{7-7}
     & Class & App. & Geo. & Spa. & Case & Obj. \\
    \midrule
    \multicolumn{1}{l}{\small\textbf{\textit{w/o \ac{llm}}}} \\
    ViL3DRel \citep{chen2022language} & 61.8 & 66.9 & 46.5 & 59.5 & 61.8 & 39.8 \\
    3D-VisTA \citep{zhu20233d} & 71.0 & 64.6 & 56.3 & 68.9 & 71.0 & 50.9 \\
    PQ3D \citep{zhu2024unifying} & \textbf{76.1} & \textbf{71.2} & \textbf{66.0} & \textbf{74.5} & \textbf{76.1} & \textbf{57.2} \\
    SceneVerse \citep{jia2024sceneverse} & 73.4 & 64.9 & 64.6 & 71.9 & 73.5 & 52.1 \\
    \midrule
    \multicolumn{1}{l}{\small\textit{\textbf{\ac{llm}-based}}} \\
    LEO-multi & 14.3 & 10.9	& 15.3 & 15.1 & 14.3 & 2.8 \\
    LEO-curricular & 22.0 & 22.2 & 20.8 & 15.4 & 22.0 & 3.8 \\
    PQ3D-LLM & 70.3 & 66.2 & 53.5 & 68.3 & 70.2 & 47.4 \\
    Chat-Scene \citep{huang2024chatscene} & 62.7 & 57.3 & 56.3 & 57.8 & 62.7 & 44.3 \\
    \bottomrule
    \end{tabular}
    }
    \label{tab:overall_results1}
\end{minipage}
\hfill
\begin{minipage}{0.52\linewidth}
    \centering
    \captionof{table}{\textbf{Evaluation results of \ac{qa} on \benchmark.} Object-centric metrics (``Obj.'') are drastically lower than case-centric metrics. $^\dagger$ indicates text input (\ie, object locations and attributes) instead of 3D point cloud.}
    \resizebox{\linewidth}{!}{
    \begin{tabular}{lccccccc}
    \toprule
     & \multicolumn{5}{c}{Knowledge type} & \multicolumn{2}{c}{Overall} \\
     \cmidrule(lr){2-6} \cmidrule{7-7} \cmidrule{8-8}
     & Class & App. & Geo. & Spa. & Exi. & Case & Obj. \\
    \midrule
    \multicolumn{1}{l}{\small\textbf{\textit{w/o \ac{llm}}}} \\
    3D-VisTA \citep{zhu20233d} & 20.5 & 33.5 & 52.1 & 33.8 & 36.5 & 35.3 & 8.1 \\
    PQ3D \citep{zhu2024unifying} & \textbf{36.4} & 28.0 & 27.8 & 11.9 & 45.5 & 27.8 & 3.5 \\
    SceneVerse \citep{jia2024sceneverse} & 35.6 & 41.7 & 48.9 & 41.9 & 35.7 & 40.3 & 6.6 \\
    \midrule
    \multicolumn{1}{l}{\small\textit{\textbf{\ac{llm}-based}}} \\
    GPT-4o$^\dagger$ \citep{openai2023gpt4} & 33.3 & \textbf{49.9} & 54.9 & \textbf{52.1} & \textbf{73.8} & \textbf{57.1} & \textbf{20.2} \\
    LEO-multi & 25.8 & 37.7 & 52.8 & 46.2 & 37.4 & 41.1 & 3.5 \\
    LEO-curricular & 17.4 & 41.0 & 53.2 & 48.7 & 39.7 & 43.2 & 7.8 \\
    PQ3D-LLM & 28.0 & 30.8 & 35.2 & 25.2 & 26.2 & 27.9 & 2.3 \\
    Chat-Scene \citep{huang2024chatscene} & \textbf{36.4} & 39.8 & \textbf{56.7} & 47.6 & 48.8 & 45.8 & 7.8 \\
    \bottomrule
    \end{tabular}
    }
    \label{tab:overall_results2}
\end{minipage}
\vspace{-10pt}
\end{table*}

\section{Experiments}\label{sec:5}

Our experiments aim to address the following questions: 
\begin{itemize}[nolistsep,noitemsep]
\item How does the object-centric evaluation scheme differ from conventional case-centric metrics in revealing model performance? (\cref{sec:exp:object_centric})
\item How do models perform when handling language variations in the \acp{gc}? (\cref{sec:exp:chain})
\item Do models show performance coherence between grounding and \ac{qa} on \acp{gqac}? (\cref{sec:exp:chain})
\item Do \acp{llm} affect the model performance? (\cref{sec:exp:llm})
\end{itemize}
To explore these questions, We select a variety of \sota \ac{3dvl} models as baselines, categorizing them based on their use of \ac{llm}. We make the necessary adjustments to ensure that most baselines can handle both grounding and \ac{qa} tasks with the same set of model weights (see implementation details in \supp). Specifically, we consider the following baseline categories in our experiments:
\begin{itemize}
\item \textbf{Without \ac{llm}.} This category includes four baselines: ViL3DRel \citep{chen2022language}, 3D-VisTA \citep{zhu20233d}, PQ3D \citep{zhu2024unifying}, and SceneVerse \citep{jia2024sceneverse}. ViL3DRel is selected as a grounding specialist and evaluated using its original checkpoint. For 3D-VisTA, we multi-task fine-tune the model to make it a generalist capable of handling both grounding and \ac{qa} tasks. For PQ3D, we directly use its pre-trained checkpoint as it is already a generalist model. For SceneVerse, we freeze the backbone pre-trained for grounding and add an additional head for fine-tuning it on the \ac{qa} task.

\item \textbf{\ac{llm}-based.} This category includes five models: GPT-4o \citep{openai2023gpt4}, LEO-multi, LEO-curricular, PQ3D-LLM, and Chat-Scene \citep{huang2024chatscene}. GPT-4o is prompted with object lists with locations and attributes for question answering. The object attributes are sourced from MSQA \cite{linghu2024multi}, which were generated using GPT-4V. LEO-multi and LEO-curricular are implemented by extending LEO \citep{huang2024embodied} to grounding through contrastive learning between object tokens and language embeddings. LEO-multi is trained with both tasks jointly while LEO-curricular is trained first on grounding and then on \ac{qa} with the backbone frozen. PQ3D-LLM is adapted from PQ3D by replacing T5-Small \citep{raffel2020exploring} with Vicuna-7B \citep{chiang2023vicuna}. Chat-Scene is evaluated directly with its checkpoint.
\end{itemize}

\begin{figure*}[t!]
    \centering
    \includegraphics[width=\linewidth]{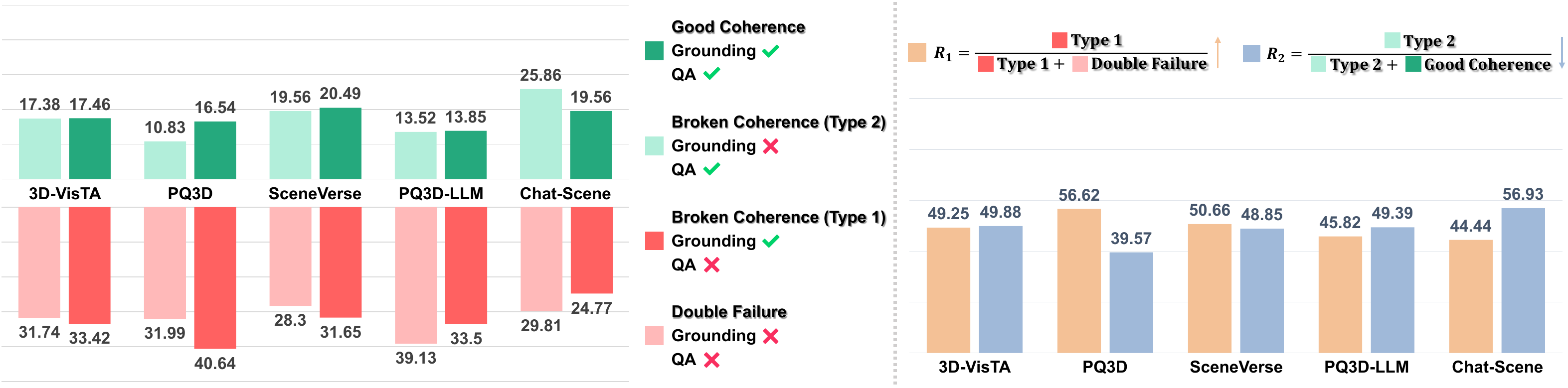}
    \caption{\textbf{Chain-of-analysis for \aclp{gqac}.} The left figure visualizes the evaluation results across \acp{gqac}, which exhibit a large proportion of broken grounding-\ac{qa} coherence. The right figure shows two metrics for evaluating broken coherence: $R_1$ for the proportion of \ac{qa} failures from insufficient \ac{qa} skills, and $R_2$ for the proportion of \ac{qa} successes from shortcuts.}
    \label{fig:gqa_analysis}
    \vspace{-12pt}
\end{figure*}

\subsection{Object-centric \textbf{\textit{vs.}} Conventional Metrics}\label{sec:exp:object_centric}
As shown in \cref{tab:overall_results1,tab:overall_results2}, we observe a significant performance drop of all \ac{3dvl} models by simply switching from per-case metrics to object-centric metrics in both grounding and \ac{qa}. In 3D grounding, we observe an average performance drop by 20\%, with LLM-based methods experiencing a more pronounced decline. For \acs{3dqa}, model performance nearly drops to zero for all models after the metric switch, except for the 2D baseline GPT-4o. These findings highlight that existing \ac{3dvl} models lack a comprehensive understanding of objects and are prone to variations in language descriptions and questions. The results underscore the importance of the object-centric evaluation scheme in pinpointing these limitations of \ac{3dvl} models. We provide additional analyses in \supp, such as discussions on outliers and the effect of \acp{llm}.

\subsection{Chain-of-analysis for Coherence Evaluation}\label{sec:exp:chain}

\begin{figure}[t!]
    \centering
    \includegraphics[width=\linewidth]{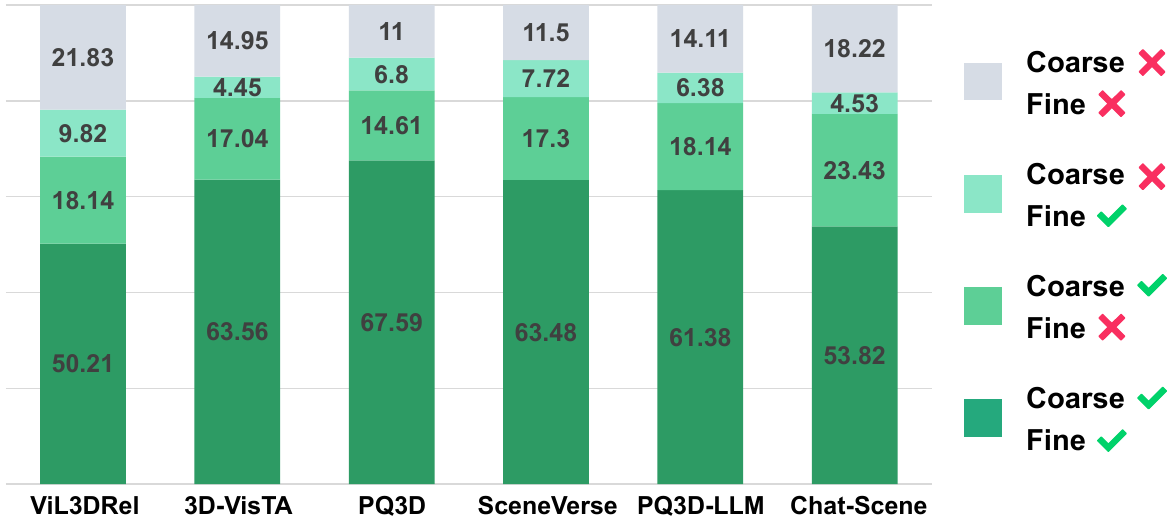}
    \caption{\textbf{Chain-of-analysis for \aclp{gc}.}}
    \label{fig:g_analysis}
    \vspace{-20pt}
\end{figure}

\paragraph{Grounding Chains.} We aggregate the evaluation results along \acp{gc} and categorize them into four types based on the grounding results on coarse (simplified grounding texts) and fine-grained texts (primary grounding texts). We leave out LEO variants in our chain analysis considering their weakness in grounding. We show the chained accuracy statistics in~\cref{fig:g_analysis}. We demonstrate that models struggle with the increased granularity in the \ac{gc}, where more failures in fine-grained primary grounding texts occur than in coarse simplified grounding texts. This indicates the difficulty of grounding primary grounding texts despite more detailed contexts, suggesting that understanding complex texts and maintaining model performance coherence across text granularities is still a challenge for \ac{3dvl} models.

\vspace{-1em}

\paragraph{Grounding-\ac{qa} Chains.} We aggregate the results across \acp{gqac} to study the gap between grounding and \ac{qa}. As shown in \cref{fig:gqa_analysis}, we categorize the results into four types based on the results of grounding and \ac{qa}. We observe a large proportion of broken coherence between tasks, echoing \cref{sec:3.3}. In particular, we design two metrics for evaluating the grounding-\ac{qa} coherence: $R_1$ for the proportion of \acp{gqac} where grounding is correct and \ac{qa} is incorrect, indicating insufficient \ac{qa} skills; $R_2$ for the proportion of \acp{gqac} where grounding is incorrect but \ac{qa} is correct, suggesting shortcuts. We find both $R_1$ and $R_2$ are close to 50\%, revealing a substantial gap between the skills of grounding and \ac{qa}, as well as the prevalence of shortcuts in \ac{qa}. This advocates deeper explorations in enhancing \ac{qa} skills and mitigating shortcuts for \ac{3dvl} models.

\subsection{Effect of \acp{llm}}\label{sec:exp:llm}

\paragraph{\acp{llm} hinder grounding.} \cref{tab:overall_results1,fig:g_analysis} show that \ac{llm}-based models perform worse than those without \ac{llm}. This includes (1) models that explicitly use \ac{llm} for grounding, such as Chat-Scene, which underperforms compared to non-\ac{llm} models like PQ3D and SceneVerse, despite excelling on existing benchmarks \citep{chen2020scanrefer,zhang2023multi3drefer}; and (2) models indirectly influenced by \ac{llm}, such as PQ3D-LLM, which performs worse than PQ3D, suggesting that integrating \ac{llm} parameters may bias the learning of grounding. These findings indicate that \ac{llm}-based models face a heightened risk of overfitting in grounding tasks.

\vspace{-1em}

\paragraph{\acp{llm} do not fundamentally enhance \ac{qa}.} While \ac{llm}-based models achieve higher per-case accuracy, this is expected given their inherent language modeling capability. However, they have not shown a fundamentally better capability in 3D QA, as evidenced by their limited accuracy in object-centric metrics (\cref{sec:exp:object_centric}) and poor grounding-QA coherence (\cref{sec:exp:chain}). This suggests that the primary bottleneck in 3D QA lies in 3D perception and \ac{vl} alignment rather than language modeling, where LLMs excel. Moreover, prior works \citep{zhu2024unifying,jia2024sceneverse} show that simple QA heads (\eg, T5-Small \citep{raffel2020exploring} and MCAN \citep{yu2019deep}) can already achieve competitive performance, indicating that 3D QA requires only basic language modeling. Therefore, improving 3D QA may depend more on advancing 3D vision foundation models than on leveraging \acp{llm}.

\subsection{Additional Insights}\label{sec:5.2.4}

\paragraph{Task.} Results in \cref{tab:overall_results1,fig:g_analysis} highlight the strong grounding capabilities of PQ3D and SceneVerse, suggesting that scaling up \ac{3dvl} data is a promising strategy for grounded 3D scene understanding. This supports training 3D vision foundation models without integrating \acp{llm}, which proves redundant and even detrimental. On the other hand, 3D QA remains highly challenging due to severe overfitting and shortcut learning in current \ac{3dvl} models. A practical solution is to start with a pre-trained backbone with strong grounding capability and then perform lightweight finetuning. This is supported by (1) SceneVerse (finetuning \ac{qa} head on top of grounding pretraining) shows best QA performances among non-LLM models, and (2) LEO-curricular (grounding-then-\ac{qa}) outperforms LEO-multi (multi-task).

\vspace{-1em}

\paragraph{Knowledge types.} We observe that geometry (Geo.) is the most challenging aspect in grounding task, probably because geometric features are rarely referenced in training data. In contrast, geometry-related questions in \ac{qa} involve less diverse answers, potentially reducing the challenge. Conversely, the diverse answers in class and appearance (App.) increase the task difficulty and lead to lower accuracy.

\section{Conclusion}

We propose \benchmark, a novel benchmark for 3D grounding and \ac{qa} tasks, delivering an evaluation paradigm shift to object-centric evaluation and analysis across grounding-\ac{qa} chains. \benchmark is driven by a detailed investigation into the limitations of existing \ac{3dvl} benchmarks, addressing flawed test data, vulnerable evaluation metrics, and the isolation of grounding and \ac{qa} tasks. Our evaluation of \sota \ac{3dvl} models highlights model pitfalls like insufficient object-level understanding, weak grounding-\ac{qa} coherence, and limited effect of \ac{llm} on \ac{3dvl} tasks.

\section*{Acknowledgments}
The authors thank Tengyu Liu for his help in setting up the annotation tool and other colleagues in BIGAI General Vision Lab for their assistance.

{
    \small
    \bibliographystyle{ieeenat_fullname}
    \bibliography{ref}
}

\appendix
\clearpage

\renewcommand{\thefigure}{A.\arabic{figure}}
\renewcommand{\thetable}{A.\arabic{table}}
\renewcommand{\theequation}{A.\arabic{equation}}
\setcounter{figure}{0}
\setcounter{table}{0}
\setcounter{equation}{0}

\section{Annotation Tool}

\begin{figure*}[t!]
    \centering
    \includegraphics[width=\linewidth]{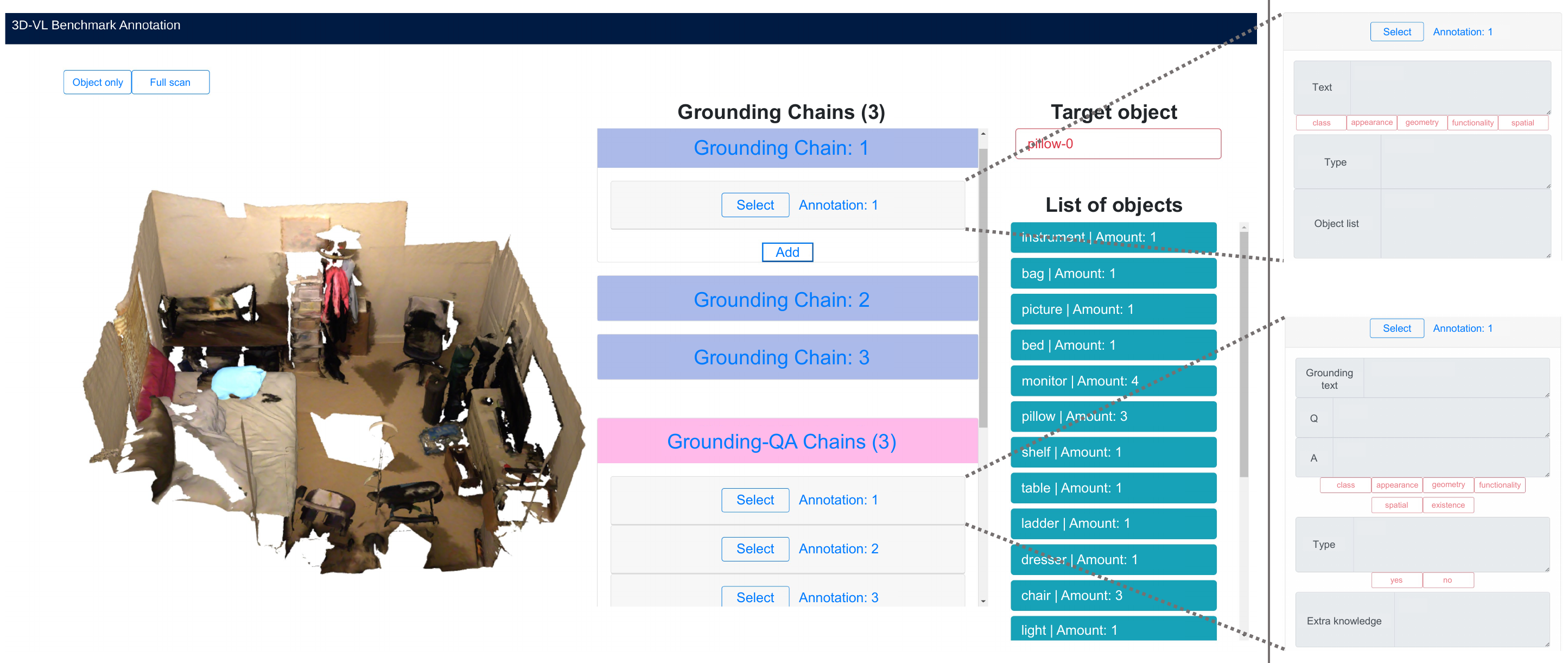}
    \caption{\textbf{Overview of our annotation tool.} The interface includes a 3D viewer (left), an annotation editor (middle), and object information (right). Two panels on the right exhibit details of each annotation for the grounding and \ac{qa} task, respectively.}
    \label{fig:ui}
\end{figure*}

We set up an interactive annotation tool for data collection based on SQA3D \citep{ma2023sqa3d}. We present a visualization of the user interface (UI) in \cref{fig:ui}, including a 3D scene viewer (left), an annotation editor (middle), and object information (right). There are three \acp{gc} and three \acp{gqac} to be annotated in the annotation editor for each target object.

Two panels on the right exhibit details of each annotation:
\begin{itemize}[leftmargin=*,label=-]
    \item For the grounding task, the human annotator is supposed to fill the referential text with precise and natural language, and then select the involved knowledge types and a list of objects that match the referential text.
    \item For the \ac{qa} task, the human annotator first generates a \ac{qa} pair based on the ``grounding text'', which lists three \textit{primary grounding texts} from the \acp{gc}. Then, the annotator labels the knowledge type and the flag of \textit{extra knowledge}, \eg, ``no'' if the answer is covered by the ``grounding text''.
\end{itemize}

\section{Baselines}\label{sec:implementation}
\paragraph{ViL3DRel \citep{chen2022language}.} This is a \ac{3dvl} specialist model for grounding, trained in a single-task scheme. We use the official checkpoint trained on ScanRefer \citep{chen2020scanrefer}.

\paragraph{3D-VisTA \citep{zhu20233d}.} While 3D-VisTA adopts task-specific fine-tuning for downstream tasks by default, we perform multi-task training by aggregating the datasets it uses. The datasets for grounding include ScanRefer, Nr3D \citep{achlioptas2020referit3d}, Sr3D \citep{achlioptas2020referit3d}, and Multi3DRefer \citep{zhang2023multi3drefer}. The datasets for \ac{qa} include ScanQA \citep{azuma2022scanqa} and SQA3D \citep{ma2023sqa3d}.

\paragraph{PQ3D \citep{zhu2024unifying}.} PQ3D is a \ac{3dvl} generalist model that supports both grounding and \ac{qa} tasks. We directly use the checkpoint after pretraining and multi-task training. The training datasets include Scan2Cap \citep{chen2021scan2cap} in addition to the datasets for 3D-VisTA.

\paragraph{SceneVerse \citep{jia2024sceneverse}.} SceneVerse is a \ac{3dvl} model pretrained on large-scale grounding datasets. To make it a generalist model for grounding and \ac{qa}, we finetune a \ac{qa} head while freezing the pretrained backbone weights to preserve its grounding ability. The datasets for fine-tuning include ScanQA and SQA3D.

\paragraph{GPT-4o \citep{openai2023gpt4}.} As a \sota \ac{llm}, GPT-4o is selected as a specialist model for \ac{qa} to probe the upper bound of \acp{llm}. We adopted the evaluation pipeline outlined in \citep{linghu2024multi} to assess GPT-4o's performance. In our evaluation, we prompt GPT-4o to answer the questions based on a collection of objects, which comprises the category, location, size, and attributes of each object. The object attributes are extracted with GPT-4V \citep{openai2023gpt4}.

\paragraph{LEO-multi.} To address the lack of grounding capability in LEO \citep{huang2024embodied}, we design a grounding loss alongside the original autoregressive language modeling loss. The grounding loss resembles contrastive learning (CLIP \citep{radford2021learning}) on the alignment between object tokens (the input to \ac{llm}) and text embeddings. With the multi-task objective, we train LEO-multi by combining grounding (ScanRefer and Nr3D) with instruction-tuning tasks (ScanQA, SQA3D, 3RScan-QA \citep{huang2024embodied}, 3RScan-Plan \citep{huang2024embodied}, and 3RScan-Dialog \citep{huang2024embodied}).

\paragraph{LEO-curricular.} Similar to LEO-multi, LEO-curricular incorporates the contrastive grounding loss but learns grounding and \ac{qa} in a curricular strategy. We first train the 3D encoder of LEO-curricular with grounding loss on ScanRefer and Nr3D. We then freeze the 3D encoder and finetune the \ac{llm} with LoRA \citep{hu2022lora} on instruction-tuning datasets.

\paragraph{PQ3D-LLM.} This is a model variant based on PQ3D, substituting the original T5-Small \citep{raffel2020exploring} with Vicuna-7B \citep{chiang2023vicuna}, which is finetuned with LoRA. The training setting is identical to PQ3D.

\paragraph{Chat-Scene \citep{huang2024chatscene}.} Chat-Scene is designed to be a \ac{3dvl} generalist model, using object identifiers and \ac{llm} to perform grounding. The training datasets include ScanRefer, Multi3DRefer, Scan2Cap, ScanQA, and SQA3D. We directly use its released checkpoint for evaluation.

\section{Additional Analyses}\label{sec:additional_analysis}

\subsection{Outliers and Prospective Questions}

We observe several outliers in our evaluation results. Below, we address these outliers and answer prospective questions:

\paragraph{\textit{Poor grounding for LEO-multi and LEO-curricular.}} The grounding performance of these two models falls significantly below that of others. We attribute this to our implementation of the grounding task learning, which employs contrastive learning between object tokens and text embeddings of pretrained \ac{llm} (\eg, Vicuna). We receive two lessons from this: (1) contrastive learning demands large-scale data while the scarce \ac{3dvl} data proves insufficient; and (2) unlike CLIP, the text embeddings of pretrained \ac{llm} may not be suitable for contrastive learning.

\paragraph{\textit{Poor \ac{qa} for PQ3D and PQ3D-LLM.}} Despite the strong performance in grounding for these two models, their performance in \ac{qa} is notably weak. We attribute this to the choice of language encoder. Compared to 3D-VisTA, PQ3D adopts a similar overall architecture but differs in language encoder: 3D-VisTA uses BERT \citep{devlin2019bert}, whereas PQ3D uses CLIP. The reasonable \ac{qa} performance of 3D-VisTA indicates that the CLIP language encoder is suboptimal for \ac{qa} task, despite being adequate for grounding. This further underscores the linguistic gap between grounding and \ac{qa} tasks: grounding texts encompass descriptive language while questions involve diverse querying patterns. It reveals the limitations of the CLIP language encoder in addressing this disparity.

\paragraph{\textit{Why is PQ3D-LLM worse than PQ3D in grounding?}} While the \ac{llm} incorporated by PQ3D-LLM is only used for \ac{qa}, it introduces a significant number of extra parameters for optimization, which may hinder the learning of grounding during multi-task learning and consequently weaken the grounding performance.

\paragraph{\textit{Why is PQ3D-LLM not better than PQ3D in \ac{qa}?}} In PQ3D, the input to the \ac{qa} head (\eg, \ac{llm}) only comprises object tokens, which can be regarded as foreign language for \ac{llm}. The challenge of utilizing these tokens for \ac{qa} cannot be alleviated by incorporating \ac{llm}, despite its strength in language processing. Additionally, incorporating \ac{llm} for \ac{qa} is prone to overfitting given the scarcity of 3D \ac{qa} data.

\paragraph{\textit{Strong performance of GPT-4o in \ac{qa}.}} We observe that GPT-4o significantly outperforms \ac{3dvl} models in \ac{qa}, especially in questions related to appearance (App.) and existence (Exi.). This showcases the upper bound of using explicit textual information (\eg, object lists with attributes), which bypasses 3D perception. The considerable gap between GPT-4o and \ac{3dvl} models further suggests that 3D perception remains a key bottleneck in \ac{3dvl} models.

\subsection{Discussion on the Effect of \ac{llm}}

\begin{table*}[t!]
\begin{minipage}{0.474\linewidth}
    \centering
    \captionof{table}{\textbf{Evaluation results of grounding on \benchmark (3RScan).} The settings and metrics follow the main paper. $^{**}$ denotes models that have never been trained in 3RScan. $^{*}$ denotes models that have been trained in 3RScan but not on grounding. $^\ddagger$ denotes only point feature is available.}
    \resizebox{\linewidth}{!}{
    \begin{tabular}{lcccccc}
    \toprule
     & \multicolumn{4}{c}{Knowledge type} & \multicolumn{2}{c}{Overall} \\
     \cmidrule(lr){2-5} \cmidrule{6-6} \cmidrule{7-7}
     & Class & App. & Geo. & Spa. & Case & Obj. \\
    \midrule
    \multicolumn{1}{l}{\small\textbf{\textit{w/o \ac{llm}}}} \\
    ViL3DRel$^{**}$ \citep{chen2022language} & 41.5 & 44.9 & 37.4 & 37.3 & 41.5 & 18.4 \\
    3D-VisTA$^{**}$ \citep{zhu20233d} & 45.6 & 38.3 & 37.4 & 40.9 & 45.6 & 21.7 \\
    PQ3D$^{**\ddagger}$ \citep{zhu2024unifying} & 38.3 & 28.0 & 36.4 & 35.3 & 38.3 & 13.6 \\
    SceneVerse \citep{jia2024sceneverse} & \textbf{61.8} & \textbf{51.4} & \textbf{53.3} & \textbf{57.3} & \textbf{61.8} & \textbf{37.5} \\
    \midrule
    \multicolumn{1}{l}{\small\textit{\textbf{\ac{llm}-based}}} \\
    LEO-multi$^{*}$ & 10.1 & 9.9 & 9.7 & 8.8 & 10.1 & 0.4 \\
    LEO-curricular$^{*}$ & 15.3 & 17.7 & 11.8 & 9.3 & 15.3 & 1.1 \\
    PQ3D-LLM$^{**\ddagger}$ & 30.3 & 27.6 & 24.6 & 25.5 & 30.3 & 8.5 \\
    \bottomrule
    \end{tabular}
    }
    \label{tab:overall_results3}
\end{minipage}
\hfill
\begin{minipage}{0.512\linewidth}
    \centering
    \captionof{table}{\textbf{Evaluation results of \ac{qa} on \benchmark (3RScan).} $^\dagger$ indicates text input (\ie, object locations and attributes) instead of 3D point cloud. $^{**}$ denotes models that have never trained in 3RScan. $^{*}$ denotes models that have been trained in 3RScan but not on \ac{qa}. $^\ddagger$ denotes only point feature is available.}
    \resizebox{\linewidth}{!}{
    \begin{tabular}{lccccccc}
    \toprule
     & \multicolumn{5}{c}{Knowledge type} & \multicolumn{2}{c}{Overall} \\
     \cmidrule(lr){2-6} \cmidrule{7-7} \cmidrule{8-8}
     & Class & App. & Geo. & Spa. & Exi. & Case & Obj. \\
    \midrule
    \multicolumn{1}{l}{\small\textbf{\textit{w/o \ac{llm}}}} \\
    3D-VisTA$^{**}$ \citep{zhu20233d} & 15.2 & 24.1 & 28.2 & 25.3 & 28.9 & 25.7 & 3.3 \\
    PQ3D$^{**\ddagger}$ \citep{zhu2024unifying} & 6.5 & 19.6 & 13.6 & 16.6 & 52.6 & 25.7 & 0.7 \\
    SceneVerse$^{*}$ \citep{jia2024sceneverse} & 28.3 & 32.3 & 34.6 & 38.9 & 44.6 & 37.4 & 0.4 \\
    \midrule
    \multicolumn{1}{l}{\small\textit{\textbf{\ac{llm}-based}}} \\
    GPT-4o$^\dagger$ \citep{openai2023gpt4} & 34.8 & 38.2 & 40.0 & 45.4 & \textbf{60.7} & \textbf{46.1} & \textbf{11.0} \\
    LEO-multi & \textbf{37.0} & 35.0 & \textbf{51.8} & \textbf{48.5} & 46.5 & 44.1 & 1.8 \\
    LEO-curricular & 19.6 & \textbf{41.8} & 48.2 & \textbf{48.5} & 50.7 & 45.6 & 7.4 \\
    PQ3D-LLM$^{**\ddagger}$ & 13.0 & 21.4 & 17.3 & 21.4 & 33.2 & 23.4 & 1.8 \\
    \bottomrule
    \end{tabular}
    }
    \label{tab:overall_results4}
\end{minipage}
\end{table*}

\paragraph{\ac{llm} hinders grounding.} This conclusion is drawn from the consideration of two categories of models:
\begin{itemize}[leftmargin=*,label=-]
    \item \textit{\ac{llm} directly used for grounding.} Models that perform grounding based on \ac{llm} (\eg, Chat-Scene) exhibit less robust performance compared to models without \ac{llm}. Specifically, despite the close performances on ScanRefer, Chat-Scene lags behind PQ3D and SceneVerse on \benchmark, which implies the potential risk of overfitting for \ac{llm}-based grounding. However, \ac{llm} may be beneficial in more complex grounding tasks that require high-level reasoning or planning, \eg, sequential grounding \citep{zhang2024task}. This suggests that the effect of \ac{llm}-based grounding varies according to task complexity.

    \item \textit{\ac{llm} not directly used for grounding.} In models that do not rely on \ac{llm} for grounding (\eg, PQ3D-LLM), we observe a weaker performance in grounding after incorporating \ac{llm}. This shows the negative effect of \ac{llm}'s parameters on the learning of grounding during multi-task learning. A practical solution is to decompose multi-task learning into curricular learning, which disregards \ac{llm}'s parameters during the learning of grounding.
\end{itemize}

\paragraph{\ac{llm} does not truly improve \ac{qa}.} We elaborate on this conclusion from three aspects: clarification on how we draw the conclusion, explanation on why per-case metrics do not matter, and analysis on why \ac{llm} may not help 3D \ac{qa}.
\begin{itemize}[leftmargin=*,label=-]
    \item \textit{How we draw the conclusion.} The evidence mainly comes from two observations: (1) the results of \ac{llm}-based models are comparable to those without \ac{llm} under object-centric metrics; and (2) fragile grounding-\ac{qa} coherence.

    \item \textit{Why per-case metrics do not matter.} While \ac{llm}-based models show slightly better results in per-case metrics, these metrics do not reliably indicate true 3D \ac{qa} capability. As demonstrated in the main paper, per-case metrics are not robust enough due to their vulnerability to shortcuts. Moreover, the advantage of \ac{llm}-based models in per-case metrics is marginal, which is intuitive given \ac{llm}'s strength in general \ac{qa}. We believe the marginal gap in per-case metrics cannot evidence a gap in the true capability of 3D \ac{qa}.

    \item \textit{Why \ac{llm} may not help 3D \ac{qa}.} We conjecture the bottleneck in 3D \ac{qa} lies in the alignment between 3D features and \ac{qa} modules, rather than language generation, where the primary strength of \ac{llm} resides. Prior works \citep{zhu20233d,zhu2024unifying,jia2024sceneverse} have shown that simple \ac{qa} heads (\eg, T5-Small or MCAN \citep{yu2019deep}) perform well in 3D \ac{qa}, as the task demands only a basic level of language generation. This explains the minimal contribution of \ac{llm} to 3D \ac{qa}.
\end{itemize}

\paragraph{Harnessing \ac{llm} for \ac{3dvl} tasks.} We first identify a critical problem in current 3D \acp{lvlm} and then propose an effective solution to harness \ac{llm} for \ac{3dvl} tasks.
\begin{itemize}[leftmargin=*,label=-]
    \item \textit{Problem.} Our investigation in the main paper reveals that overfitting to text is a critical problem in current 3D \acp{lvlm}. This implies a significant imbalance between 3D encoder and \ac{llm}, that is, \ac{llm} often overshadows 3D encoder during training. This issue is less pronounced in 2D \acp{lvlm} owing to the robust 2D features learned through extensive pretraining, which is infeasible for 3D encoders.

    \item \textit{Solution.} We propose curricular learning, progressing from grounding to \ac{qa}, as an effective solution to mitigate this issue by shielding 3D features from \ac{llm} interference. The effectiveness is evidenced by the advantages of SceneVerse and LEO-curricular.
\end{itemize}

\subsection{Limitations and Future Work}
First, our benchmark prioritizes focused and systematic analysis, which involves trade-offs in task scope and complexity. Our object-centric evaluation excludes more advanced tasks, such as multi-object grounding and complex reasoning. Extending this evaluation framework to include more complex tasks will be a key direction for future work. Second, our baselines may not cover the wide range of existing \ac{3dvl} models. We will evaluate and analyze more models in the future. Third, we consider the performance of the grounding task as a proxy for the grounding implicitly performed in the \ac{qa} task. This may be unfair to models whose grounding performance is locked due to issues like improper implementation (\eg, LEO-multi and LEO-curricular). Nonetheless, we believe our approach remains practical for assessing grounding-\ac{qa} coherence in most \ac{3dvl} generalist models.

\section{Domain Transfer}

\begin{table*}[t!]
\begin{minipage}{0.47\linewidth}
    \centering
    \captionof{table}{\textbf{Evaluation results of grounding on \benchmark (MultiScan).} The settings and metrics follow the main paper. $^{**}$ denotes models that have never been trained in MultiScan. Only SceneVerse has been trained in MultiScan.}
    \resizebox{\linewidth}{!}{
    \begin{tabular}{lcccccc}
    \toprule
     & \multicolumn{4}{c}{Knowledge type} & \multicolumn{2}{c}{Overall} \\
     \cmidrule(lr){2-5} \cmidrule{6-6} \cmidrule{7-7}
     & Class & App. & Geo. & Spa. & Case & Obj. \\
    \midrule
    \multicolumn{1}{l}{\small\textbf{\textit{w/o \ac{llm}}}} \\
    ViL3DRel$^{**}$ \citep{chen2022language} & 33.2 & 34.4 & 25.0 & 32.0 & 33.2 & 13.2 \\
    3D-VisTA$^{**}$ \citep{zhu20233d} & 40.8 & 30.5 & 28.1 & 38.0 & 40.8 & 18.9 \\
    PQ3D$^{**}$ \citep{zhu2024unifying} & 56.3 & 53.9 & 37.5 & 52.8 & 56.3 & 34.0 \\
    SceneVerse \citep{jia2024sceneverse} & \textbf{59.5} & \textbf{54.6} & \textbf{53.1} & \textbf{56.6} & \textbf{59.5} & \textbf{35.9} \\
    \midrule
    \multicolumn{1}{l}{\small\textit{\textbf{\ac{llm}-based}}} \\
    LEO-multi$^{**}$ & 9.0 & 9.1 & 9.4 & 9.0 & 9.0 & 1.3 \\
    LEO-curricular$^{**}$ & 11.7 & 11.0 & 6.3 & 9.0 & 11.7 & 0 \\
    PQ3D-LLM$^{**}$ & 51.0 & 46.8 & 37.5 & 49.0 & 51.0 & 25.8 \\
    \bottomrule
    \end{tabular}
    }
    \label{tab:overall_results5}
\end{minipage}
\hfill
\begin{minipage}{0.52\linewidth}
    \centering
    \captionof{table}{\textbf{Evaluation results of \ac{qa} on \benchmark (MultiScan).} $^\dagger$ indicates text input (\ie, object locations and attributes) instead of 3D point cloud. $^{**}$ denotes models that have never been trained in MultiScan. $^{*}$ denotes models that have been trained in MultiScan but not on \ac{qa}.}
    \resizebox{\linewidth}{!}{
    \begin{tabular}{lccccccc}
    \toprule
     & \multicolumn{5}{c}{Knowledge type} & \multicolumn{2}{c}{Overall} \\
     \cmidrule(lr){2-6} \cmidrule{7-7} \cmidrule{8-8}
     & Class & App. & Geo. & Spa. & Exi. & Case & Obj. \\
    \midrule
    \multicolumn{1}{l}{\small\textbf{\textit{w/o \ac{llm}}}} \\
    3D-VisTA$^{**}$ \citep{zhu20233d} & 6.5 & 22.6 & 16.7 & 13.2 & 28.8 & 19.1 & 0 \\
    PQ3D$^{**}$ \citep{zhu2024unifying} & 21.0 & 16.8 & 16.7 & 9.6 & 39.0 & 20.8 & 0.6 \\
    SceneVerse$^{*}$ \citep{jia2024sceneverse} & 16.2 & 32.1 & 12.5 & 26.5 & 38.1 & 28.9 & 3.1 \\
    \midrule
    \multicolumn{1}{l}{\small\textit{\textbf{\ac{llm}-based}}} \\
    GPT-4o$^\dagger$ \citep{openai2023gpt4} & \textbf{29.0} & \textbf{41.6} & 33.3 & 25.7 & \textbf{59.3} & \textbf{39.4} & \textbf{7.6} \\
    LEO-multi$^{**}$ & 12.9 & 24.1 & 41.7 & 24.3 & 32.2 & 25.6 & 2.5 \\
    LEO-curricular$^{**}$ & 8.1 & 27.0 & \textbf{50.0} & \textbf{28.7} & 41.5 & 29.8 & 3.8 \\
    PQ3D-LLM$^{**}$ & 6.5 & 21.9 & 8.3 & 11.0 & 25.4 & 17.0 & 0.6\\
    \bottomrule
    \end{tabular}
    }
    \label{tab:overall_results6}
\end{minipage}
\end{table*}

We follow the setting outlined in the main paper to evaluate the baselines in two novel domains: 3RScan \citep{wald2019rio} and MultiScan \citep{mao2022multiscan}. This evaluation is referred to as \textit{domain transfer} since most baselines are only trained on ScanNet \citep{dai2017scannet}. Notably, as Chat-Scene only provides model features for ScanNet, its evaluation on 3RScan and MultiScan is not feasible. We distinguish between two types of domain transfer:
\begin{itemize}[leftmargin=*,label=-]
    \item $^{**}$: the model has never been trained in the target domain.
    \item $^{*}$: the model has been trained in the target domain but on tasks other than the specific one.
\end{itemize}

\paragraph{Results.} We present the domain transfer results for 3RScan in \cref{tab:overall_results3,tab:overall_results4}, and MultiScan in \cref{tab:overall_results5,tab:overall_results6}. The overall trends are consistent with those reported in the main paper for ScanNet. For example, while models without \ac{llm} (\eg, SceneVerse) excel in grounding, \ac{llm}-based models (\eg, LEO-curricular) perform better under per-case metrics but struggle with object-centric metrics in \ac{qa}. In particular, we report several specific findings regarding the domain transfer results:
\begin{itemize}[leftmargin=*,label=-]
    \item \textbf{\textit{Challenge of domain transfer.}} All models exhibit notable performance declines, emphasizing the challenge of domain transfer (ScanNet $\rightarrow$ 3RScan; MultiScan). SceneVerse surpasses PQ3D owing to its comprehensive pretraining across diverse scene domains. Moreover, training on 3RScan-QA improves \ac{qa} performance on 3RScan (LEO-multi and LEO-curricular). These findings highlight the inevitable domain gap and the benefits of cross-domain pretraining.
    
    \item \textbf{\textit{Limitations of feature-dependent models.}} PQ3D and PQ3D-LLM experience considerable performance drops on 3RScan due to a lack of image and voxel features. While this issue results in only a marginal drop on ScanNet, as reported in the original paper \citep{zhu2024unifying}, the considerable drop on 3RScan indicates the heightened challenges of transferring to novel domains for feature-dependent models such as PQ3D and Chat-Scene.
    
    \item \textbf{\textit{More challenging 3D perception in MultiScan.}} Performance on MultiScan is consistently lower than on 3RScan, reflecting the increased difficulty of 3D perception in the domain of MultiScan. SceneVerse, despite using a simple \ac{qa} head \citep{yu2019deep}, outperforms LEO-multi and matches LEO-curricular. This suggests that the bottleneck in \ac{qa} lies in 3D perception, suppressing the contribution of \ac{llm}. It further underscores the need for more powerful 3D encoders to address this bottleneck.
    
    \item \textbf{\textit{Performance degradation of GPT-4o.}} GPT-4o exhibits noticeably lower performance on 3RScan and MultiScan compared to ScanNet, with the results on 3RScan approached by LEO-curricular. We attribute this degradation to incomplete object attributes stemming from insufficient multi-view images, which limits the object attribute extraction by GPT-4V. This reveals that, despite their strengths in 3D \ac{qa}, \acp{llm} and 2D \ac{lvlm} are constrained by the availability of high-quality multi-view images.
\end{itemize}

\section{Illustration of Data and Evaluation}

\begin{figure*}[t!]
    \centering
    \includegraphics[width=\linewidth]{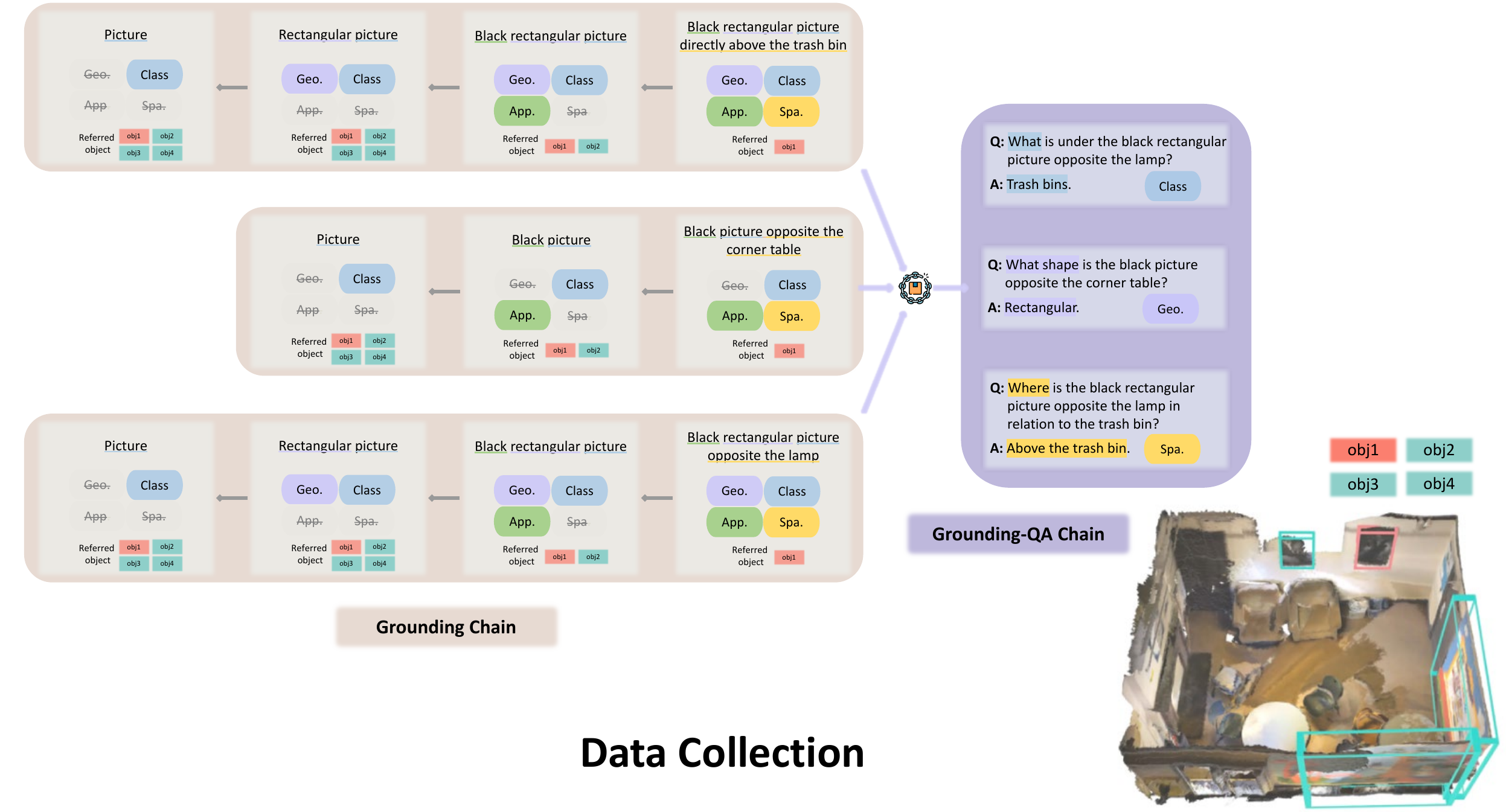}
    \caption{\textbf{Static overview of data collection.} Check the dynamic process in our video demo in the attachment.}
    \label{fig:data_vis}
\end{figure*}

\begin{figure*}[t!]
    \centering
    \includegraphics[width=0.8\linewidth]{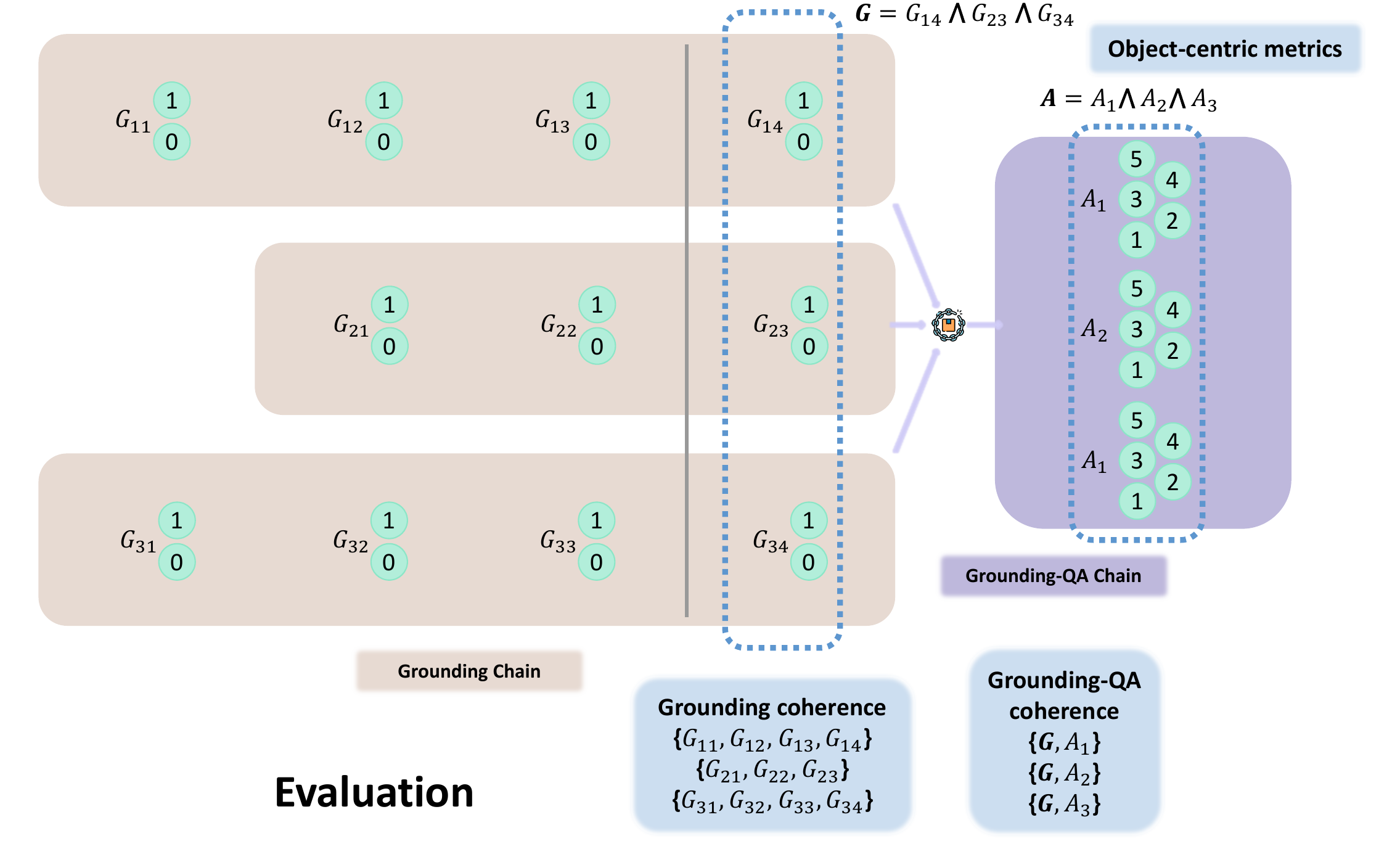}
    \caption{\textbf{Static overview of evaluation.} Check the dynamic process in our video demo in the attachment.}
    \label{fig:eval_vis}
\end{figure*}

We present a video demo to illustrate the process of data collection and evaluation (see attachment). Here we show the static overview in \cref{fig:data_vis,fig:eval_vis}.

\end{document}